%% file: main_paper.tex
\documentclass[manuscript]{acmart}

\usepackage{cleveref}
\usepackage{booktabs} % For formal tables

\usepackage{algorithm2e}
\usepackage{algorithm2e}

\usepackage{xcolor}
\usepackage{graphicx} 
\usepackage{lipsum}   
\usepackage{caption}
\usepackage{array} 
\usepackage{graphicx} 
\usepackage{subcaption} 
\usepackage{listings}
\usepackage{multirow}
\usepackage{amsmath}
\usepackage{algorithm}
\usepackage{algpseudocode}

\definecolor{darkred}{RGB}{139,0,0}

\definecolor{zyd_1}{HTML}{FF8899} 
\definecolor{zyd_2}{HTML}{77BBDD}

\newcommand{\onedot}{.}
\newcommand{\eg}{\textit{e.g.}}
\newcommand{\etc}{\textit{etc.}}
 
\def\ie{\emph{i.e}\onedot}

\SetAlFnt{\small}
\SetAlCapFnt{\small}
\SetAlCapNameFnt{\small}
\SetAlCapHSkip{0pt}

\begin{document}
% \begin{document}
\sloppy

\title{VC-Agent: An Interactive Agent for Customized Video Dataset Collection}

\author{Yidan Zhang}
\orcid{0000-0002-0082-1920}
\affiliation{%
 \institution{SSE, The Chinese University of Hong Kong, Shenzhen}
 \streetaddress{Longxiang Ave 2001}
 \city{Shenzhen}
 \state{Guangdong}
 \postcode{518172}
 \country{China}}
\email{yidanzhang@link.cuhk.edu.cn}

\author{Mutian Xu}
\orcid{0000-0001-8123-6493}
\authornote{corresponding author}

\affiliation{%
 \institution{SSE, The Chinese University of Hong Kong, Shenzhen}
 \streetaddress{Longxiang Ave 2001}
 \city{Shenzhen}
 \state{Guangdong}
 \postcode{518172}
 \country{China}}
\email{mutianxu@link.cuhk.edu.cn}

\author{Yiming Hao}
\orcid{0009-0000-8185-4400}
 \affiliation{%
 \institution{FNii-Shenzhen}
 \streetaddress{Longxiang Ave 2001}
 \city{Shenzhen}
 \state{Guangdong}
 \postcode{518172}
 \country{China}}
\affiliation{%
 \institution{SSE, The Chinese University of Hong Kong, Shenzhen}
 \streetaddress{Longxiang Ave 2001}
 \city{Shenzhen}
 \state{Guangdong}
 \postcode{518172}
 \country{China}}
\email{haoym1016@gmail.com}

\author{Kun Zhou}
\orcid{0000-0001-9592-6575}
\affiliation{%
 \institution{SSE, The Chinese University of Hong Kong, Shenzhen}
 \streetaddress{Longxiang Ave 2001}
 \city{Shenzhen}
 \state{Guangdong}
 \postcode{518172}
 \country{China}}
\email{zhoukun303808@gmail.com}

\author{Jiahao Chang}
\orcid{0009-0009-6877-1649}
\affiliation{%
 \institution{SSE, The Chinese University of Hong Kong, Shenzhen}
 \streetaddress{Longxiang Ave 2001}
 \city{Shenzhen}
 \state{Guangdong}
 \postcode{518172}
 \country{China}}
\email{224010128@link.cuhk.edu.cn}

\author{Xiaoqiang Liu}
\orcid{}
\affiliation{%
 \institution{Kuaishou Technology}
 \streetaddress{Building 1, 6 Shangdi West Road}
 \city{Beijing}
 \state{Beijing}
 \postcode{100085}
 \country{China}}
\email{liuxiaoqiang@kuaishou.com}

\author{Pengfei Wan}
\orcid{}
\affiliation{%
 \institution{Kuaishou Technology}
 \streetaddress{Building 1, 6 Shangdi West Road}
 \city{Beijing}
 \state{Beijing}
 \postcode{100085}
 \country{China}}
\email{wanpengfei@kuaishou.com}

\author{Hongbo Fu}
\orcid{}
\affiliation{%
 \institution{The Hong Kong University of Science and Technology}
 \streetaddress{Clear Water Bay}
 \city{Hong Kong}
 \state{Hong Kong}
 \postcode{}
 \country{China}}
\email{hongbofu@ust.hk}

\author{Xiaoguang Han}
\orcid{0000-0003-0162-3296}
\affiliation{%
 \institution{SSE, The Chinese University of Hong Kong, Shenzhen}
 \streetaddress{Longxiang Ave 2001}
 \city{Shenzhen}
 \state{Guangdong}
 \postcode{518172}
 \country{China}}

 \affiliation{%
 \institution{FNii-Shenzhen}
 \streetaddress{Longxiang Ave 2001}
 \city{Shenzhen}
 \state{Guangdong}
 \postcode{518172}
 \country{China}}
\email{hanxiaoguang@cuhk.edu.cn}

\input{sec/0_abstract}

\begin{teaserfigure}
    \centering
    \includegraphics[width=1.0\linewidth]{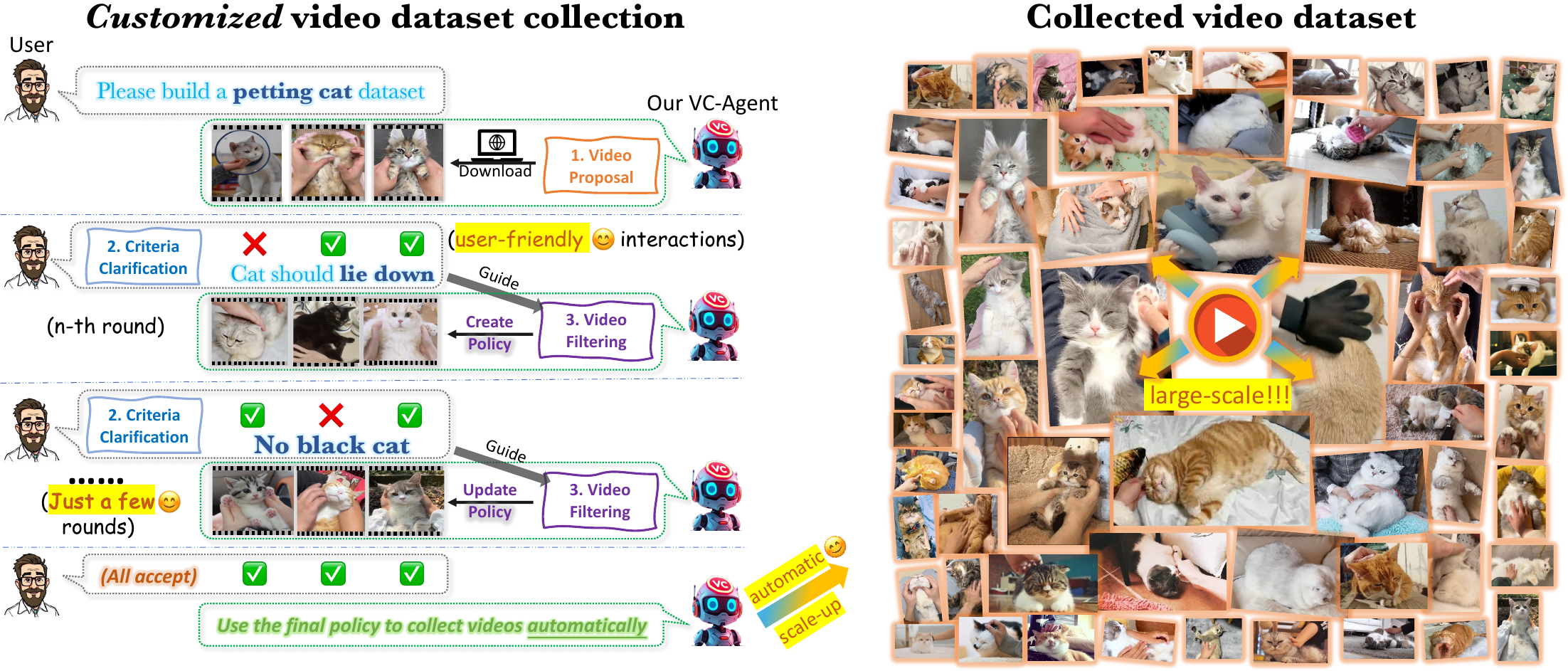}
    \caption{\textit{We propose \textbf{VC-Agent}, the first \textbf{interactive} MLLM-based agent that can effectively scale-up the collection of \textbf{customized} video datasets from Internet}.
    With a few \textit{iterative interactions} requiring \textit{minimal} user input, both the user's demands and our filtering policy evolve progressively towards the optimal goal of a customized dataset. After several rounds of interactions, our agent utilizes the final updated policy to collect videos automatically.
    } 
    \label{fig:teaser}
\end{teaserfigure}


\maketitle

\input{sec/1_intro}

\input{sec/2_relatedWork}
\input{sec/3_methology}
\input{sec/4_experiments}
\input{sec/5_user_study}
\input{sec/7_conclusion}

\bibliographystyle{ACM-Reference-Format}
\bibliography{main_paper}

\input{sec/X_suppl}

\end{document}

%% file: sec/0_abstract.tex
\begin{abstract}
Facing scaling laws, video data from the internet becomes increasingly important. However, collecting extensive videos that meet specific needs is extremely labor-intensive and time-consuming. In this work, we study the way to expedite this collection process and propose VC-Agent, the first interactive agent that is able to understand users' queries and feedback, and accordingly retrieve/scale up relevant video clips with minimal user input. Specifically, considering the user interface, our agent defines various user-friendly ways for the user to specify requirements based on textual descriptions and confirmations. As for agent functions, we leverage existing multi-modal large language models to connect the user's requirements with the video content. More importantly, we propose two novel filtering policies that can be updated when user interaction is continually performed. Finally, we provide a new benchmark for personalized video dataset collection, and carefully conduct the user study to verify our agent's usage in various real scenarios. Extensive experiments demonstrate the effectiveness and efficiency of our agent for customized video dataset collection. Project page: https://allenyidan.github.io/vcagent\_page/.

\end{abstract}

%% file: sec/1_intro.tex
\section{Introduction} 

Video datasets play a crucial role in advancing deep learning techniques for video-related tasks, such as video understanding \cite{laurenccon2024obelics,schuhmann2021laion400mopendatasetclipfiltered}, video generation \cite{ho2022imagen, yang2023diffusion} and 3D reconstruction \cite{mildenhall2020nerfrepresentingscenesneural,kerbl20233dgaussiansplattingrealtime}. In particular, web-scale video data from the Internet, being one of the most natural and easily accessible sources of video data, has brought a transformative impact to the community. By training on vast amounts of Internet video data, recent video generative models \cite{zhou2022magicvideo,villegas2022phenaki, zhao2023survey, li2023videogen,wu2023tune, kondratyuk2023videopoet, zhang2024show,bar2024lumiere,gupta2025photorealistic} have demonstrated astonishing creativity and generalization capabilities.

Despite the abundant video data available on the internet, manually selecting qualified videos to meet specific needs and constructing a \textit{customized} video dataset from them is a laborious and time-consuming task for a human collector \cite{xu2018youtube,miao2021vspw,miao2022large,huang2024wildavatar}, making it hard to efficiently scale up personalized data collection.
To address this, we aim to \textit{automate} the customized video data collection.
Thanks to recent large models for language, visual and cross-modal understanding \cite{guo2025deepseek, achiam2023gpt,touvron2023llama,brown2020language,zhu2023minigpt,liu2024visual,qwen2}, an intuitive baseline is to use a website video crawler to grasp videos based on summarized text descriptions about users' requirements and
employ Multimodal Large Language Models (MLLMs) to retrieve relevant visual content from videos according to these summaries. 
% \todo{(draw a baseline figure? showing very difficult and bad results)}. 
However, this approach has the following problems: 1) It is commonly challenging for users to describe their requirements using text all at once, without browsing video samples.
2) Even if the requirements can be summarized, they tend to be very complex and detailed, posing significant challenges for models to understand, even with the aid of current large models.

Inspired by the observation that human collectors usually need to engage in \textit{multiple} rounds of communication with users to align their requirements, in this work, we aim to design a framework to tackle the aforementioned challenges.
Taking a closer look at the general procedure for constructing the customized video dataset, as \cref{fig:teaser} (left) shows, it can be divided into three stages. 1) \textit{Video proposals}: Beginning with a rough requirement from the user, the collector downloads relevant videos from the Internet; 2) \textit{Criteria clarification:} According to video proposals, users further clarify more requirements.
% that prompt the collector to conduct video filtering.
3) \textit{Video filtering}: Based on user demands, the collector filters videos that clearly do not meet requirements, and keeps videos that meet the criteria.

The last two stages are typically conducted in an \textit{iterative} manner, during which the newly filtered videos will be reviewed by the user again, leading the criteria to grow \textit{increasingly specific/certain}, thereby guiding the filtering to become \textit{more accurate} at the same time. 
After several iterations, the user will be satisfied with all filtered videos, prompting the collector to finalize the collection by filtering videos according to previously summarized demands.

This procedure can be further disentangled from the viewpoint of user interface (\textit{front}-end) and agent functions (\textit{back}-end). On one hand, regarding \textbf{user interface}, our agent supports three communication ways: 1) \textit{initial query-} At the beginning, users can provide a rough text to outline the desired characteristics of the target video. 2) \textit{confirmations-} Following the initial query, the agent can propose numerous candidate video clips. Among them, users can then confirm the videos that meet their requirements. 3) \textit{comments-} For candidate videos that are unsatisfactory, users can provide some comments to describe the reasons. On the other hand, to fulfill users' demands, two critical \textbf{agent functions} need to be realized: 1) The agent should take a textual description as input and \textit{propose} candidate video clips that are as relevant as possible. 2) The agent should possess a \textit{policy} capable of determining acceptance or rejection for any input video clip. In particular, the policy should be \textit{updatable} based on received confirmations and comments.

To this end, we introduce a novel framework, called \textbf{\textit{VC-Agent}}, the first interactive MLLM-based agent that can efficiently scale up the customized video dataset collection. At the front end, our agent calls the \textit{User Interface} module (\cref{sec:user_interface}) to actively interact with users. This involves prompting the user to input the initial query and provide feedback by reviewing the retrieved video samples for confirmation and comments.
At the back end, to realize the aforementioned agent functions, we first carefully develop a \textit{Video Proposal} module (\cref{sec:video_proposal}) that combines MLLMs and video grounding models to understand the user's demands and extract a pool of candidate video clips from Internet. More importantly, we novelly define and formulate two updatable policies for accepting and rejecting candidate videos via interacting with users (\cref{sec:filtering_policy}). Specifically, we propose a \textit{Template-Based Acceptance Policy}, where several videos that have been confirmed by users are gathered and summarized by MLLMs, to create positive criterion templates. 
Moreover, we design an \textit{Attribute-Aware Rejection Policy}, where the user's comments of rejection reasons are summarized by MLLMs specifically according to several attributes, forming a negative standard table.
% (\eg, appearance -- not black; pose -- standing; \etc).
Based on positive criterion templates and the negative standard table, MLLMs are utilized again to describe and filter other candidate videos.
\textit{Notably}, in the next iterations, by reviewing the newly filtered videos, the user becomes clearer about the requirements, providing newly confirmed videos and rejection reasons that serve as new sources to update the filtering policy continually. This \textit{virtuous cycle} progressively refines \textit{\textbf{both user demands and the filtering policy}} to achieve the optimal goal of a customized dataset.
In addition, to enhance the robustness of our system, we introduce a double-check strategy that prompts the user to review videos filtered by MLLMs with low confidence scores.
After a few iterations with minimal user input, our agent transitions into a fully automatic mode to scale up the dataset without requiring user interaction.

To verify the effectiveness of our agent, we provide a new benchmark for personalized video dataset collection, where 10,000 Internet videos are annotated according to specific and distinct requirements. Furthermore,  we carefully conduct the user study to verify our agent's usage in various real scenarios.

Our contributions are as follows: 
\begin{itemize}
    \item We propose VC-Agent, the first interactive agent that efficiently scales up the customized video dataset construction with minimal user input.
    \item We carefully design a novel pipeline by leveraging existing large language models, including a framework for proposing video clip candidates and an iterative video filtering mechanism with two updatable policies.
    \item We develop an easy-to-use Web UI for untrained users to utilize our agent.
    \item We provide a new personalized video collection benchmark, including 10,000 videos with extensive annotations of three distinct domains of personalized requirements.
    \item Extensive experiments verify that our agent can effectively collect the customized video dataset, greatly reducing users' time while ensuring quality. On average, each dataset only requires minutes for interaction. 
    \item We will release the code, Web UI, benchmark, and datasets produced by our agent, scaling from 5k to 80k.
\end{itemize}

%% file: sec/2_relatedWork.tex
\section{Related Work}
\subsection{(Building) Video Dataset}
A key driver behind the success of video-related tasks is the availability of large-scale video datasets. For instance, in the action recognition domain, datasets have grown from early collections like HMDB-51, with 3,312 videos \cite{kuehne2011hmdb}, and UCF-101, with 2,500 videos \cite{soomro2012ucf101}, to much larger datasets such as Kinetics-700, with 306,245 videos \cite{kay2017kinetics}. However, assembling these datasets often involves downloading videos from online sources and manually filtering them, which is time-consuming and costly. This process becomes even more challenging as datasets expand into specialized domains \cite{huang2024wildavatar, xiong2024mvhumannet, ng2022animal, chen2023mammalnet, yu2023mvimgnet, das2013thousand, zhou2018towards} with strict selection criteria. 

To overcome these challenges and facilitate the flexible collection of large-scale video datasets, we introduce a novel interactive system to reduce dataset construction costs and adapt readily to various specialized video tasks.

\subsection{Video Retrieval}
Video retrieval, aiming to find relevant video content based on user queries, has gained significant attention as video data grows rapidly across various fields. Video retrieval methods can be divided into visual-based, text-based, and hybrid models. Visual-based methods~\cite{kordopatis2019visil,jo2022exploring} often use similarities between video frames to retrieve videos that match query images,  

while text-based video retrieval approaches~\cite{fang2021clip2video,portillo2021straightforward} apply diverse text/video embedding techniques to achieve text-video alignment. Recently, pre-trained Vision-Language Models (VLMs) have shown substantial potential in understanding video and multi-modal tasks. For instance, generative models for text/image/video-to-video~\cite{wang2024recipe,zhoustructure,gupta2025photorealistic,hu2024animate} demonstrate the feasibility of more flexible and accurate multi-modal video retrieval.

Video retrieval systems rely on single-modality input (\eg, text-only), posing challenges for precise requirement specification. Instead, our method defines \textit{diverse} interaction schemes for \textit{precise} demand expression, which are \textit{compatibly} supported by our policy design, enabling our agent to retrieve target videos accurately.

\subsection{MLLM Agent}
An AI agent is an autonomous entity capable of making decisions and taking actions based on its real-time environment \cite{wooldridge1995intelligent}. With advancements in Large Language Models (LLMs), notably their impressive reasoning and comprehension abilities, researchers have begun to explore their applications across various tasks \cite{Han_Liu_Wang_2023, liu2024visual, ye2023mplug, yang2023mm}. This has led to the development of AI agents based on LLMs. LLM-agent can not only communicate naturally with users but also leverage exceptional comprehension and generalization capabilities to address complex real-world challenges \cite{wei2022chain}, such as decision-making \cite{yang2024doraemongpt}, robot navigation \cite{brohan2023rt}, tool utilization \cite{shen2024hugginggpt}, and collaborative model integration \cite{yang2024v}. Recently, LLMs and VLMs have converged with each other, giving rise to Multimodal Large Language Models (MLLMs). MLLMs use LLMs as the brain to receive, process, and generate outputs based on multimodal information \cite{yin2024survey}. Particularly in video understanding tasks, researchers have enabled MLLM-agents to achieve remarkable performance \cite{fan2025videoagent, wang2024videoagent, sleimangoldfish} and extended them through tool integration \cite{gao2023assistgpt, suris2023vipergpt}. 

In this work, we develop an interactive system that strategically integrates MLLMs with a self-updatable policy for backend decision-making, enabling efficient and scalable construction of customized video datasets.

%% file: sec/3_methology.tex
\section{VC-Agent}

\cref{fig:framework} illustrates the framework of our proposed VC-Agent, which highlights the interaction between the user and agent.
In this section, we describe the details of our method from the perspective of the user interface (\textit{front}-end) and agent functions (\textit{back}-end), respectively.
On one hand, as for the user interface, \cref{sec:user_interface} (\textbf{User Interface}) describes several basic ways of how the user interacts with the agent to express requirements.
On the other hand, there are two main agent functions.
1) In \cref{sec:video_proposal} (\textbf{Video Proposal}): according to the user's requirements, VC-Agent retrieves raw video clips from public platforms and applies a video grounding model to select the most relevant clips from the initial collection.
2) In \cref{sec:filtering_policy} (\textbf{Filtering Policy}): we introduce an innovative filtering policy that allows VC-Agent to iteratively refine its process, ensuring an optimal match with the user's specifications.

\input{img_table_tex/fig_pipeline}
\subsection{User Interface}
\label{sec:user_interface}

User interface is the front end of our agent, facilitating a comprehensive understanding of the user's requirements. It consists of three basic user interactions: initial query, confirmation, and comment.

\paragraph{Initial query.}
At the start of the dataset construction, the user initiates with a coarse query $Q$, such as “Please build a petting cat dataset” as shown in \cref{fig:framework}. The agent will automatically fetch relevant data from video platforms, retrieving and filtering videos.

\paragraph{Confirmation.}
Multiple samples will be randomly selected from filtered videos and sent back to the user. Subsequently, the user will be tasked to choose acceptable and unacceptable videos, leading to the formation of two distinct sets: the acceptance set $V_+ = { v_1, v_2, \ldots, v_n}$ and the rejection set $V_- = { v_1, v_2, \ldots, v_n}$.
The accepted videos will serve as reference templates for updating the acceptance policy, while the rejected ones will help the user to understand and update their rejected criteria.

\paragraph{Comment.}
\label{sec:comment}

Moreover, users can provide reasons for rejection to form the comment set $C = \{c_{v} \mid v\in {V_p}\}$. Specifically, as illustrated in \cref{fig:framework}, the user may specify preferences such as ``No black cat" or ``Cat should lie down" on rejected videos. Both $V_-$ and $C$ will play a pivotal role in updating the rejection policy.
As for accepted videos, since comments on them are usually less specific (\eg, “the video looks good”) and provide limited information, the Comment will only be called for rejected ones. The accepted videos $V_+$ will be directly utilized to enhance the acceptance policy.

The Confirmation and Comment will be iteratively invoked within the User Interface until no videos are rejected. Finally, our agent will proceed to the fully automated stage to build the dataset.

\begin{figure*}[t]
    \centering
    \includegraphics[width=1.0\textwidth]{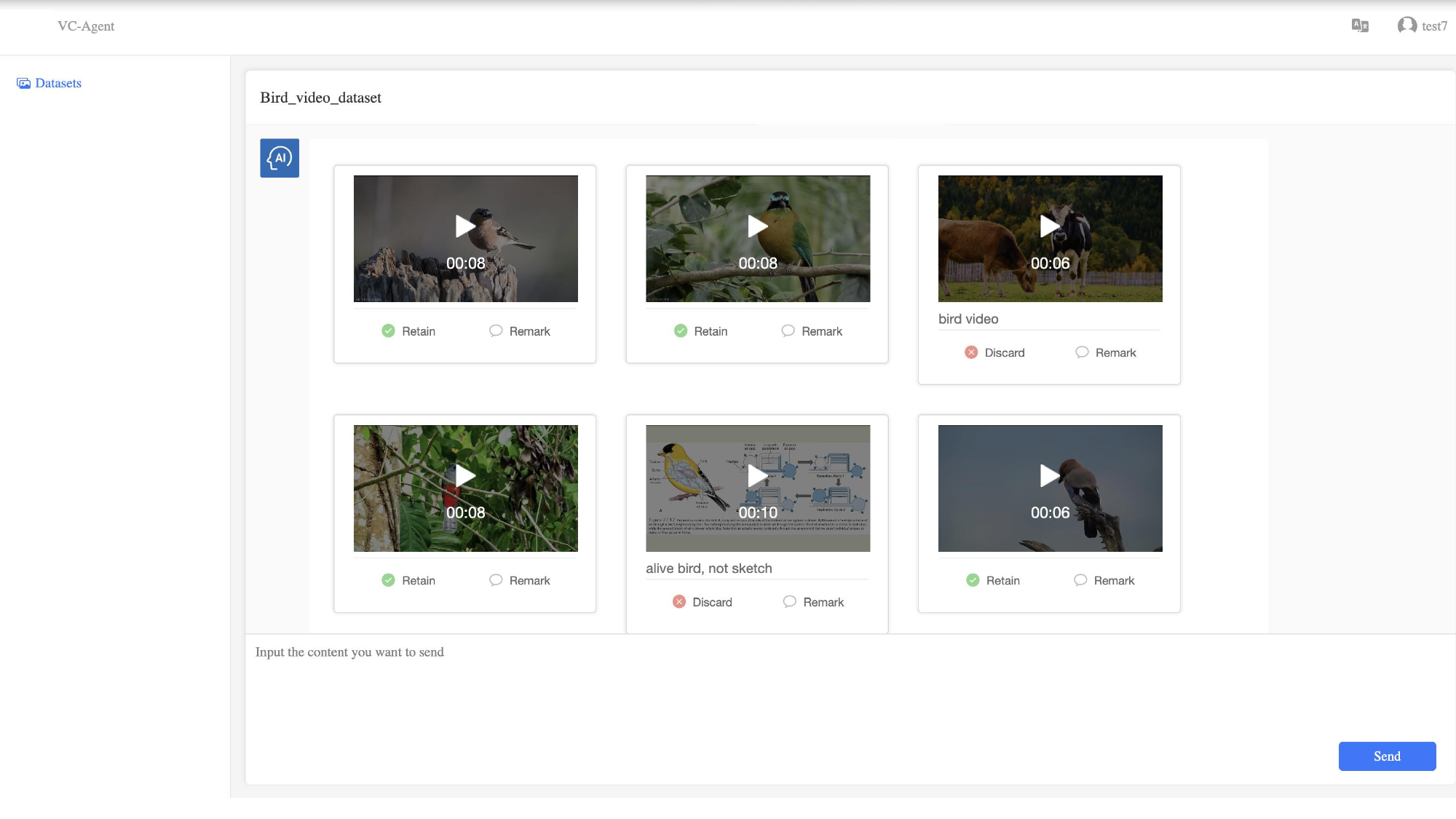}
    \caption{Our web-based user interface. The user can type the query in the bottom chatbox and click “send”, prompting our agent to perform video retrieval and return video samples to the center window. Next, the user can choose “retain” or “discard” and provide additional comments on the returned videos.}
    \label{fig:web_ui}
\end{figure*}

\paragraph{Web UI}
We develop a trivial-to-use Web UI to easily operate the aforementioned interfaces for untrained users. As shown in \cref{fig:web_ui}, the user can type the query in the bottom chatbox and click “send”, prompting our agent to perform video retrieval and return video samples to the center window. Next, the user can choose “retain” or “discard” and provide additional comments on the returned videos. The flow of data through the system is handled by a tracking server. More details and the demo video of Web UI can be found in the supplementary material.

\subsection{Agent Functions}
\label{sec:agent_function}
The back end of our agent is constructed by realizing two substantial agent functions that are based on MLLMs, defining how to retrieve and filter videos, and serving as the core of our VC-Agent. 
The prompts to MLLMs are listed in the supplementary material.

\subsubsection{Video Proposal}
\label{sec:video_proposal}

As for agent functions, the first is to propose \textit{candidate} videos according to the user's initial requirement.

\paragraph{Retrieval.}
Upon receiving the initial query $Q$ from the user, VC-Agent generates the query keyword based on the LLM module. With keywords $K$, the agent will manage the website crawler to automatically retrieve and download related videos from the public video platform to generate the video set $V_s = \{ v_1, v_2, \ldots, v_n \}$.

\paragraph{Video grounding.}
As the retrieved $V_s$ often contains a significant amount of irrelevant content at both temporal and spatial levels, we use the video grounding model to identify and extract the segments most relevant to the user query $Q$. This process involves: $V_g = f_g(V_s, Q)$, where $f_g$ represents the video grounding model.

To effectively process in-the-wild videos, the grounding model must exhibit strong capabilities in handling various videos with out-of-distribution content. To achieve this, we employ TFVTG \cite{zheng2025training} as our temporal grounding module, leveraging its superior generalization ability, which is based on the video-language model (VLM) of BLIP-2 Q-Former \cite{li2023blip}. For spatial grounding, we integrate Grounding DINO \cite{liu2023grounding}, an open-source model renowned for its exceptional efficacy.

% \vspace{-0.2cm}
\paragraph{Candidate initialization.}
Given the video clips $V_g$, our agent invokes MLLM to describe the content of $V_g$, and calculates their similarity with the user demand. The videos with $Top_K$ similarities will be initialized as candidate video proposals $V_p$. This process is formulated as:
\begin{equation}
\setlength{\abovedisplayskip}{5pt}
\setlength{\belowdisplayskip}{5pt}
    V_p = Top_K\{S(v,LLM(Q)), v \in V_g \}
\label{eq:candidate}
\end{equation}
where $S(.)$ is the similarity function, $LLM(Q)$ means the user query summarized by LLM.
As for $S(.)$, we have tried to compare text embedding similarities, or directly use LLM to rank, and both perform well.

\subsubsection{Filtering Policy}
\label{sec:filtering_policy}
Given candidate videos, the core of our agent lies in how to filter videos based on the user's feedback. Here, we propose an Attribute-Aware Rejection Policy and a Template-Based Acceptance Policy, which are dynamically updated in each iteration.

The rationale for using two distinct policies stems from the observed differences in how the user defines and comments on accepted versus rejected videos, which are detailed below.

\paragraph{Attribute-Aware Rejection Policy.}
For rejected videos, the user's feedback often highlights \textit{\textbf{specific attributes}}. For example, the user may comment, “I do not want a \textit{black} cat” in reference to \textit{appearance}. If a new video shares this attribute, it will be rejected and discarded. This insight leads us to propose an attribute-aware rejection policy.

1) \textit{Getting negative standard table from the user's comments}: In our pipeline, several candidate videos $V_r$ sampled from $V_p$ will be sent back to the user for reviewing, confirmation, and providing comments on rejected videos. Then our agent will collect the user's comments $C$ of rejection reasons on each video in the rejection set $V_-$, and utilize LLM to summarize user-concerned \textit{attributes} as well as the corresponding demands from each comment, forming a \textit{negative “standard table”}.
For example, as shown in \cref{fig:framework}, after receiving the user feedback $C$, the LLM module summarizes the attributes of concern to the user (\eg, Appearance, Action) and their corresponding values (\eg, black, standing), and stores them in the standard table.

2) \textit{Filtering videos according to standard table}: During the filtering, for a candidate video from $V_p$, our agent uses MLLM to analyze and describe the video content based on the existing attributes in the standard table. This design drives MLLM to focus on each attribute, producing an \textit{attribute-specific description} of a video. Then, this description will be sent to LLM to compare with the corresponding values (\ie, demands) of that attribute recorded in the standard table. If a video's description is found to be similar, this video will be discarded. For example, for the new batch of videos, our agent uses MLLM to examine the character's appearance and action in each video, compares these attributes with the values recorded in the dictionary, and decides whether to discard or retain the video. The process can be formulated as:
\begin{equation}
\setlength{\abovedisplayskip}{5pt}
\setlength{\belowdisplayskip}{5pt}
    V = \{ v \mid \neg S(V_p,D_a(v_i)), \forall a \in A_- , \exists v_i \in V_-  \}
    \label{eq:reje}
\end{equation}
where $S(.)$ is the Similarity Function, $D(.)$ denotes the standard table, $a$ means one of the negative attributes $A_{-}$ summarized from comments $C$, $V_p$ is candidate videos.

\paragraph{Template-Based Acceptance Policy.}
In contrast to the rejection policy, the user generally can not provide specific reasons to accept a video, but can only confirm it. This means that the whole information of an accepted video should be considered as a global acceptance criterion, and different videos may share \textit{similar} criteria to decide acceptance.
This underlying logic encourages us to design a template-based acceptance policy.

1) \textit{Generating positive criterion templates from accepted videos}: In our framework, after confirming the sampled candidate videos by the user, an acceptance set $V_+$ will be generated. $V_+$ will then be analyzed and described by MLLM. Next, our agent will send these positive descriptions to LLM, asking LLM to gather (\ie, aggregate) and summarize a few description templates.

2) \textit{Filtering videos based on positive templates}:
After the candidate videos are filtered by the rejection policy, our agent will employ LLM to explicitly compare the descriptions of the remaining videos with the positive description templates, and keep similar videos.

\paragraph{Iterative policy update.} 
% Update method
\textit{Notably}, in the next iterations, a new batch of videos $V$ will be selected from newly filtered videos and sent back to the user again. Through the review of new video samples, the user becomes clearer on their requirements, offering newly confirmed videos $V_+$, $V_-$ and comments $C$ that serve as new sources for continuously updating the acceptance policy (\ie, criterion templates) and rejection policy (\ie, standard table). This virtuous cycle incrementally refines both user demands and our filtering policy toward the optimal expectation of a customized dataset.

Furthermore, to prevent a scenario (although never appears in our practice) in which all initially sampled candidates are accepted—thereby providing no guidance for refining the rejection policy—we enforce a minimum number of interaction rounds to ensure adequate policy optimization.
If no negative samples emerge after multiple rounds, this indicates that the user’s requirements are simple, and the optimized \textit{acceptance} policy can reliably process them.

\paragraph{User-assisted double check.}
\input{img_table_tex/fig_doublecheck}
The aforementioned acceptance and rejection policies will generate a similarity score to guide the filtering process. This score reflects the confidence level of our policy in its decision-making. For instance, cats with mixed black and white colors may receive a similarity score of 40$\sim$60\%, indicating an ambiguous decision.

To enhance the robustness of our framework, we introduce a double-check strategy. Samples with low confidence are retained in a dedicated buffer. Once 100 such samples have been accumulated, a subset is randomly selected for user review during the interaction phase, prompting the user to verify them again.
Moreover, since the low-confidence samples filtered by our rejection policy are associated with specific ambiguous attributes, our agent explicitly guides the user’s attention to those aspects. For example, it will prompt: “Please double-check whether the color meets the requirements for these samples and provide additional feedback.”, as shown in \cref{fig:doublecheck}.
The resulting detailed user comments (\eg, “No \textit{purely} black”) are then used to refine the rejection policy. Similarly, samples that are confirmed by the user during double-checking are used to update and improve the acceptance policy.

This strategy maximizes the utilization and effectiveness of user interactions. Through additional user engagement without altering the original user interface, it significantly enhances the overall system's accuracy and robustness (see \cref{tb:vf_abl}).

\paragraph{Final automatic scale-up.}
Once the user remains consistently satisfied with all returned videos over multiple rounds, our agent will employ the finalized policy to seamlessly shift into a fully automated mode. This eliminates further need for human interaction, enabling efficient and scalable dataset construction.

%% file: img_table_tex/fig_pipeline.tex
\begin{figure*}[t]
    \centering
    \includegraphics[width=0.9\textwidth]{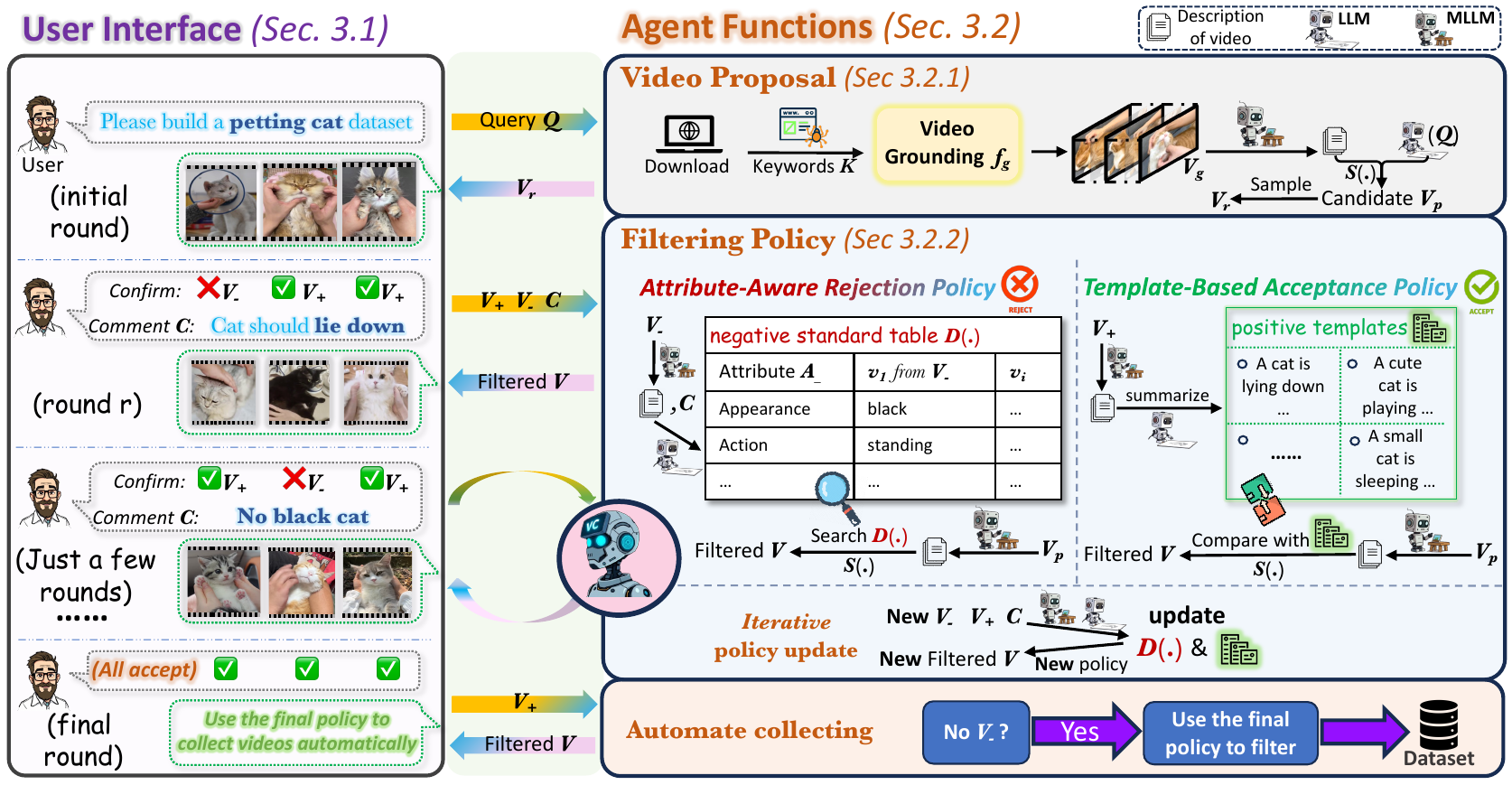}
    \caption{The illustration of the entire workflow of our VC-Agent. It mainly consists of User Interface (\textit{front}-end, \cref{sec:user_interface}, shown in the left panel) and Agent Functions (\textit{back}-end, \cref{sec:agent_function}, shown in the right panel).
    Initially, the user interacts with the agent to express their coarse demands $Q$. Upon receiving $Q$, our agent controls the Video Proposal module (\cref{sec:video_proposal}) to download and retrieve candidate videos. Next, several candidates $V_r$ will be sent back to the user, requiring them to provide the accepted/rejected video set $V_+$/$V_-$, and comment set $C$. Subsequently, $V_+$, $V_-$ and $C$ will guide ourn agent to define/update the Filtering Policy (\cref{sec:filtering_policy}). The updated policy is then used to filter candidate videos, after which a new batch of video samples $V$ will be returned to the user again for the next interaction iteration. The whole process will be \textit{iteratively} conducted until the user is satisfied with all returned videos. Finally, our agent will begin to construct the dataset in a fully automatic manner. }
    \label{fig:framework}
\end{figure*}

%% file: img_table_tex/fig_doublecheck.tex
\begin{figure*}[t]
    \centering
    \includegraphics[width=0.7\textwidth]{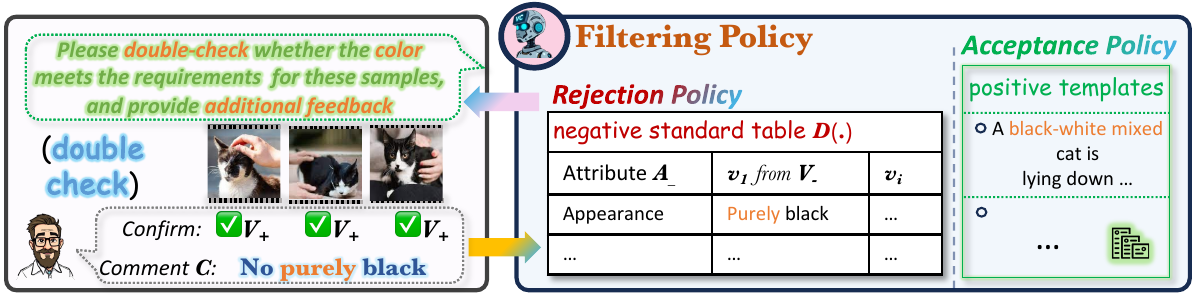}
    \caption{The overview of our user-assisted double-check strategy.}
    \label{fig:doublecheck}
\end{figure*}

%% file: sec/4_experiments.tex
\section{Main Experiments}

\subsection{Benchmarking Customized Video Collection}

\paragraph{The importance and necessity of a new benchmark.}
To evaluate the effectiveness of VC-Agent, one possible way is to use the benchmark of traditional video retrieval tasks \cite{chen2023vast,wang2024internvideo2,cicchetti2024gramian}. However, this is infeasible when evaluating our task:

1) The task targets are \textit{fundamentally different}.
The traditional video retrieval task focuses on \textit{analyzing the relevance} between a video and a \textit{single coarse} query prompt in a \textit{pre-defined} query space. Instead, our task aims to \textit{extract} videos according to \textit{iteratively updated} user demands, from \textit{web-scale} Internet videos.
2) Consequently, from the agent's perspective, solely accepting simple keywords or phrases as used in traditional video retrieval tasks makes it hard to retrieve the videos that satisfy the user's personalized demands.
Inversely, from the user's viewpoint, the user's demand may not only concern the overall content of the video but also encompass specific local attributes, such as appearance, object number, pose, \etc, which are not aligned with the target of traditional video retrieval tasks.

To remedy the defect of the benchmark, we propose a new \textbf{P}ersonalized \textbf{V}ideo Collection \textbf{B}enchmark (PVB). It contains \textbf{three} distinct styles/domains of personalized video requirements.
\input{img_table_tex/tab_res_compare}
\paragraph{Personalized Video Collection Benchmark.}

1) \textit{Search videos and provide requirements}. 
For all three domains, we totally search and download 10,000 videos according to the query keywords from Internet. Following a meticulous examination of these videos, for each domain, we precisely outline five specific requirements from five distinct perspectives.

Specifically, for the first domain:
\begin{footnotesize}
\begin{itemize}
    \item R.1. Category: Bird video.
    \item R.2. Action: Perching on the tree.
    \item R.3. Shot: Close-up shot.
    \item R.4. Component: Bird cannot be obscured by the subtitles.
    \item R.5. Content: No human in the video.
\end{itemize}
\end{footnotesize}

Next, for the second domain:
\begin{footnotesize}
\begin{itemize}
    \item R.1. Category: Cat video.
    \item R.2. Action: Lying Down.
    \item R.3. Appearance: No Black Cat.
    \item R.4. Shot: Close-up shot.
    \item R.5. Content: Single Cat.
\end{itemize}
\end{footnotesize}

Last, for the third domain:
\begin{footnotesize}
\begin{itemize}
    \item R.1. Category: Dancing video.
    \item R.2. Shot: Close-up shot.
    \item R.3. Style: Realistic Video / Not Animation
    \item R.4. Content: Single Person Dance.
    \item R.5. Content: Chinese-Style Dressing Garment.
    % \item R.5. Film Content: 
\end{itemize}
\end{footnotesize}

More \textit{benchmark statistics} such as specific video amounts are presented in the supplementary material.

2) \textit{Annotate videos according to requirements}.
We employ workers to \textit{manually} review and annotate all videos, labeling whether each video meets each five requirements or not, separately. In addition, another group of workers will double-check to make sure the annotation is accurate.

3) \textit{Key metrics}. We use the Intersection-over-Union ($IoU=\frac{TP}{TP+FP+FN}$) to measure the similarity between the retrieved video subset and the ground truth subset that meets requirements.

NOTE: During our evaluation, videos are defined as “\textit{high-quality videos}” when their \textit{semantic contents} satisfy the requirements, and pay less attention to traditional technical concepts such as temporal completeness, frame stability, or audio fidelity.

\paragraph{Experimental setups.}

1) \textit{Methods in comparison}. On one hand, we aim to compare our method with traditional video retrieval methods. We employ the widely-used InternVideo-L \cite{wang2024internvideo2}, and two recent state-of-the-art models, VAST \cite{chen2023vast} and GRAM \cite{cicchetti2024gramian}. On the other hand, we also evaluate several state-of-the-art open-sourced MLLMs, including MiniGpt4-Video \cite{ataallah2024minigpt4}, LLAVA-Next-Video-7B \cite{zhang2024llava} and LLAVA-OneVision-7B \cite{li2024llava}.

2) \textit{Implementations}. 
Since directly inputting all requirements at once to the model may bring large ambiguity, and it is especially \textit{in}feasible to evaluate the actual performance of video retrieval methods that can only accept a single query or limited input tokens.
To this end, we utilize an \textit{iterative} evaluation scheme.
Specifically, we first employ our agent and other methods to select videos according to the first requirement R.1, and conduct the evaluation. At the next iteration, all models are performed again on the \textit{previously collected} videos of R.1, to evaluate if the selected videos meet R.2, indicated by R.1\&2. The process is iteratively conducted until all requirements are evaluated. This follows the original split of the benchmark, and provides a unified scheme to evaluate the actual model performance.

In addition, as video retrieval models select the videos according to specific relevance thresholds, we report the best results by traversing various thresholds. 
All the other configurations follow the original models.
For our agent, we use Qwen2.5-7B \cite{qwen2.5} as LLM and LLAVA-OneVision-7B \cite{li2024llava} as MLLM.

\paragraph{Experimental results.}
As shown in \cref{tb:result_vf}, our VC-Agent outperforms all baseline methods.
Notably, when more requirements are introduced (\eg, from R.1\&2\&3 to R.1\&2\&3\&4\&5), the performance gap becomes larger.

This suggests that: 1) As the complexity of requirements increases, traditional video retrieval models struggle to accommodate the multitude of specific demands simultaneously. 
2) As the criteria delineated from R.1 to R.5 increasingly emphasize finer details, our agent can better handle such detailed demands.
The samples collected by our agent and other methods are compared in \cref{fig:abl_effect} and the supplementary material.

\input{img_table_tex/fig_abl_effect}

In terms of time efficiency, leveraging MLLMs incurs a higher computational cost compared to traditional video retrieval methods (1.53s per video in our approach vs. 0.42s for video retrieval). However, our method achieves significantly better performance and generalization capability.

\subsection{Ablation Studies}
We explore the impact of key modules in our method, including: 1) Removing the rejection policy; 2) Removing the acceptance policy; 3) Removing the template-based designs from the acceptance policy, \ie, comparing candidate videos with the \textit{whole} accepted video set without constructing templates; 4) Removing the attribute-aware methods from the rejection policy, \ie, summarizing rejection reasons or comparing candidate videos \textit{without} splitting attributes; 5) Removing the iterative updates in our agent by only using the first round $V_r$ to initialize the policy.

The results in \cref{tb:vf_abl} demonstrate the effectiveness of each module.
 
As the number of requirements increases, the performance gain from each module becomes more significant.

\input{img_table_tex/tab_ablation}

%% file: img_table_tex/tab_res_compare.tex
\begin{table}[t]

\resizebox{0.7\linewidth}{!}{
\centering
\begin{tabular}{l| l l l l l l }
\hline
Type&Requirements   &R.1        & R.1\&2        & R.1\&2\&3     & R.1\&2\&3\&4  &R.1\&2\&3\&4\&5  \\ \hline
VR &InternVideo-L   & 18.03   & 12.00     & 9.65    & 7.08    & 5.06    \\

&VAST                      & 21.12   & 13.19     & 11.98    & 7.87    & 5.47          \\
&GRAM            & 21.33   & 13.66     & 12.29    & 7.31    & 5.53     \\ \hline
MLLM&MiniGpt4-Video      & 25.31    &21.88    & 18.99   & 16.52     & 14.65   \\
&LLAVA-Next-Video-7B     & 48.39    &40.04   & 35.84   & 30.41     &30.12               \\
&LLAVA-OneVision-7B       & 50.42    &41.61   &36.72  &32.82 & 31.07            \\
 \hline
&\textbf{Ours}                       &\textbf{64.82}  & \textbf{60.58}       & \textbf{56.23}  & \textbf{52.95} & \textbf{49.17} \\ \hline
\end{tabular}

}

\caption{The quantitative comparison of video retrieval methods (VR), MLLMs, and our VC-Agent on Personalized Video Collection Benchmark. R.n indicates the n-th requirement that the video needs to meet. R.1\&2 denotes both R.1 and R.2 need to be satisfied, and so on. The results are calculated by the average IoU (\%) across three domains in the benchmark.
}
\label{tb:result_vf}
% \vspace{-0.8cm}
\end{table}

%% file: img_table_tex/fig_abl_effect.tex
\begin{figure}[t]
    \centering
    \includegraphics[width=1.0\linewidth]{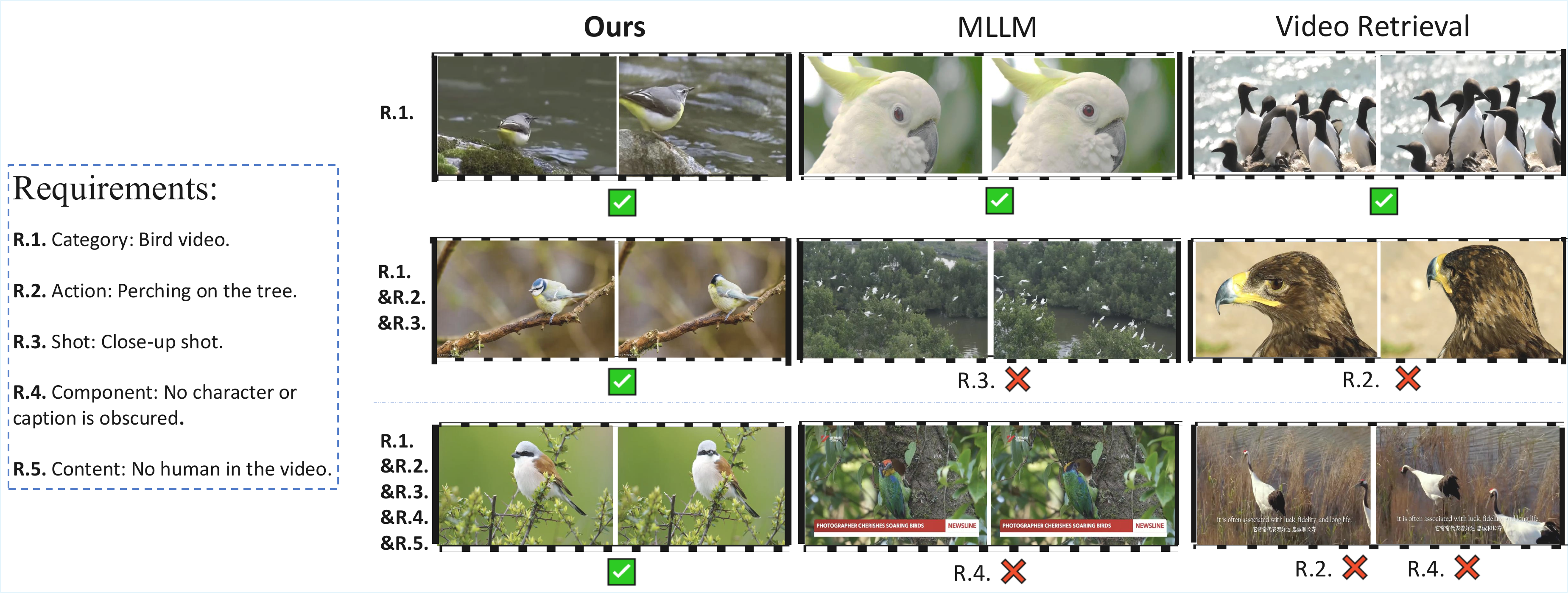} % 替换 为你的图片文件名
    \caption{The progressive filtering results obtained by incrementally adding different requirements. “Ours” denotes filtering conducted using the VC-Agent, “MLLM” refers to LLAVA-OneVision \cite{li2024llava}, and “Video Retrieval” represents GRAM \cite{cicchetti2024gramian}. As the requirements increase, our agent consistently maintains strong performance, whereas both MLLM and Video Retrieval methods gradually exhibit errors. The requirements for which filtering errors occurred are indicated below the images.}
    \label{fig:abl_effect}
\end{figure}

%% file: img_table_tex/tab_ablation.tex
\begin{table}[t]
\resizebox{0.4\linewidth}{!}{
\centering
\begin{tabular}{l l l l }
\hline
Requirements   & R.1  & R.1\&2\&3 & R.1\&2\&3\&4\&5  \\ \hline
w/o rej             &56.32   &48.15      & 36.41  \\ 
w/o acc             &60.02   &53.72      & 44.90 \\ 
w/o att in rej      &58.44   &49.20      & 39.16 \\
w/o tmp in acc      &62.20   &54.15      & 46.52  \\ 
w/o update          &59.62   &51.52      & 39.40 \\ 
w/o double-check    &59.84   &49.03      & 43.05           \\ \hline
\textbf{VC-Agent (Ours)}           & \textbf{64.82}     & \textbf{56.23}         & \textbf{49.17}   \\ \hline
\end{tabular}
}
\caption{Quantitative ablations. The results are calculated by the average IoU (\%) across different domains in the benchmark.}
\label{tb:vf_abl}
\end{table}

%% file: sec/5_user_study.tex
\section{User Study}

\subsection{Preparation}
\paragraph{Background.}
The ultimate goal of our agent is to satisfy various \textit{real-world} demands of customized video dataset constructions. To evaluate the usability and effectiveness of VC-Agent, we carefully designed user studies and made a significant effort to conduct them, including a usability study and a follow-up survey. 

\paragraph{Participants.} 
We hired eight participants (five men and three women) who had experience in collecting datasets. Among them, six are graduate students with backgrounds in computer vision, and two are algorithm engineers from the Information Technology industry. To begin with, we spent $\sim$10 minutes (average time on each participant) to introduce the operation of our framework, especially on User Interface, ensuring participants get familiar with it.

\input{img_table_tex/tab_us_statistics}

\subsection{Usability Study}
\paragraph{Setups.}
In the usability experiment, we invite participants to use our agent to construct a video dataset based on their backgrounds and actual demands. 
After extensive usage of our agent, we ask participants to complete the questionnaire to assess the progress and quality of the dataset construction one week later.

\paragraph{Questionnaire and metrics.}
We design a comprehensive questionnaire to understand users' experiences with our VC-Agent. It is divided into five parts:

1) \textit{General information}: including the expected name of datasets, users' purpose for constructing the dataset, and the amount of data collected. Notably, we do not purposely select common classes but adhere to original user requirements (\eg, biped cartoon categories are uncommon).

2) \textit{The details of the usage process}: investigating how many rounds interacted with our agent, the number of times the system initiates the user-assisted double-check process, and the total time spent to collect the data (including both user interaction and final automated processing).

3) \textit{The quality of the collected data}: we ask users to randomly sample 100 videos from the dataset collected by our agent, review each one, and calculate the ratio that meets their requirements. To reduce variance and enhance reliability, this process is repeated five times. In each evaluation iteration, a new set of 100 samples is randomly selected from the remaining unreviewed data. The final results are reported as both the accumulated and average satisfaction rates with variance.

4) \textit{The estimated time of manual collection}: users are asked to estimate how much time it would take to collect a similarly sized video dataset based on their previous experiences. 
Notably, it is not based on users' guess, but derived rigorously: users are asked to manually collect 50$\sim$100 videos \textit{continuously} and record time, which is scaled to match the final dataset size collected using our agent.
Given that manual large-scale collection requires substantially more time, the estimation deviations become statistically negligible.

5) Additionally, users' feedback and suggestions are collected and provided in the supplementary material.

\paragraph{Results.}
As \cref{tab:collected_data} shows, eight participants collected 335,428 video data points using our framework over 486 hours (690 / hour).
Our agent significantly reduces users' estimated time for the manual data collection, and almost all participants expressed high satisfaction with the data quality. Note that we asked users to evaluate output videos \textit{as strictly as possible} $\rightarrow$ many failed cases are \textit{borderline acceptable} (\eg, VC-BiCar videos showing cartoon bodies but missing one hand). Consequently, the reported data quality is not always superior (\ie, $>$ 90\%).
Nevertheless, these “borderline” cases introduce minimal noise to downstream training, potentially improving model robustness.

Moreover, we observed that approximately 10 interaction rounds—typically including 1 to 3 double-check rounds—are generally sufficient to construct a high-quality dataset. This demonstrates that our agent operates effectively with minimal user input.
More ablations are provided in the supplementary material.

In addition, we conduct \textit{System Usability Scale} (SUS) -- a widely used questionnaire scheme designed to evaluate the usability of a system, product, or service. The details and results are presented in the supplementary material.

\subsection{Follow-up Survey}
We conducted a follow-up survey with two participants who utilized our agent in the previous usability study.

In this survey, we aim to further examine \textit{detailed feedback} from users, especially on the \textbf{\textit{actual benefit}} from the data collected using our agent, which mainly focuses on \textbf{\textit{two important downstream tasks}} -- text-to-video generation and biped cartoon characters pose estimation.

% \vspace{-0.2cm}
\paragraph{Benefit on text-to-video generation.}
The user found that existing text-to-video (T2V) generative models often struggle under \textit{specialized/customized} text inputs.
To address this, the user employed our agent to collect a substantial amount of videos by providing specific requirements for a customized sub-domain (\ie, bird and feline).
Subsequently, the user finetuned a state-of-the-art T2V model, CogVideoX \cite{yang2024cogvideox, hong2022cogvideo}, using 20K data collected by our agent.
To compare with our method, the user also utilized LLAVA-OneVision-7B \cite{li2024llava} to conduct the same process.
The specific text inputs (\ie, requirements) are presented in the supplementary material.

\cref{tab:tv_res} shows the quantitative comparison collected from the user, where four commonly used T2V metrics (clip \cite{hessel2021clipscore}, hpsv2 \cite{wu2023human}, tf and sc \cite{huang2024vbench}) are reported. The model finetuned using the collected data from our agent gets better performance, while achieving more realistic and containing more details, as illustrated by the qualitative results in \cref{fig:t2v_us2} and \cref{fig:t2v_extra}.
Instead, the model finetuned using the data gained from LLAVA-OneVision slightly degrades the original performance, potentially due to the poor data quality.

\paragraph{Benefit on biped cartoon characters pose estimation.}
The user hopes to perform the pose estimation on biped cartoon characters, and contribute to various applications such as character design and motion capture. This task aims to estimate 13 key feature points of biped humanoid animated characters to determine the overall pose of the animated figure. However, current pose estimation methods often fail on this target, as they primarily focus on authentic humans.
To tackle this, the user utilized our agent to collect 61,972 animated film videos within three days, including the bipedal characters. Next, they annotated over 10,000 poses of animated characters, which are used to finetune the Rtmpose-l \cite{jiang2023rtmpose}. Requirements are presented in the supplementary material.

The quantitative results are collected from the user and presented in \cref{tab:pose_res}. As a result, using the data collected by our agent to finetune the model significantly boosts the performance. Furthermore, according to the qualitative results shown in \cref{fig:sub}, the features of animated characters can be effectively identified by the finetuned model. For instance, the rabbit ears are positioned at the top of the head rather than on the sides. In addition, the finetuned model can accurately estimate the pose with an occlusion between characters, such as a beaver's arm being obstructed by the rabbit's body.

\input{img_table_tex/tab_us_res_quant}
\input{img_table_tex/fig_us_visual_sample}

\paragraph{Reflections on these two surveys: “general VS specific”.}
As reflected from these two surveys, the baseline models that are trained with a \textit{large-scale/diverse/general-purpose dataset} frequently \textit{underperform} in specialized domains, and fine-tuning by customized data significantly improves. This underscores the need for specific-domain datasets — precisely the focus of our work.
Furthermore, our data framework can still satisfy \textit{diverse} applications (as demonstrated in other user studies).

\input{img_table_tex/fig_abl_rounds}

\paragraph{More user effort, better performance?}
A key property of an interactive system is whether expending more user effort leads to better performance.
To investigate this more clearly, we evaluate two variables related to user effort: the number of interaction rounds and the number of samples reviewed per round.
We instructed users of VC-T2V and VC-BiCar to collect an additional 1000 videos by conducting interactions across (3, 5, 7, 10, 13) rounds, or by reviewing (3, 5, 7, 9, 11) samples per round.
The results are presented in \cref{fig:abl_round}. Performance bottlenecks can be observed both in terms of the number of rounds and the number of samples per round, offering practical guidance for achieving an optimal balance between accuracy and user effort. A potential explanation is that as user effort increases, their requirements tend to become more complex, which in turn makes it more challenging for MLLMs to accurately interpret and fulfill them.

%% file: img_table_tex/tab_us_statistics.tex
\begin{table*}[t]
\centering
\resizebox{1.0\textwidth}{!}{
\begin{tabular}{lllccccc}
\hline
DatasetName     & Video Amount & Purpose Task                      & Round      &Double-Check     & Data Quality (Avg.)  & Est. Manual Time   & Actual Time (h) \\ \hline
VC-T2V           & 82,119       & Text-to-Video Generation          &   13      &2          & 86.6\%$\pm${0.4}  &  3-4 weeks           & 101    \\
VC-BiCar         & 61,972       & Biped Cartoon Pose Estimation     &   12      &2          & 84.2\%$\pm${1.2}  &  10 days             & 93    \\
VC-Sport          & 53,391       & Sport Video Dataset              &   7       &3          & 91.4\%$\pm${1.6}  &  3 weeks               & 86    \\
VC-YouCook       & 14,689       & Cooking Instruction Dataset       &   10      &1          & 87.8\%$\pm${0.8}  &  20 days    & 24    \\
VC-EgoStreetView        &  84,146      & First-view Street View Dataset &  12   &3          & 90.6\%$\pm${0.4}  &  1 month     &  122    \\

VC-Scene         & 5,210        & Indoor Scene Understanding        &   8       &3          & 86.2\%$\pm${0.6} &  7-9 days        & 10    \\
VC-Driving        &  14,907      & First-view Driving Video Dataset &   10      &2          & 92.4\%$\pm${1.4}  &  14 days      & 22   \\
VC-Dance            & 18,994       & Human Dance Pose Estimation     &   7      &1          & 88.4\%$\pm${1.0} &  2 weeks     & 28   \\
Total               & 335,428     & -           & -                            & -         & - & -              & 486    \\ \hline
\end{tabular}%
}

\caption{The results of our questionnaire, which investigates the fundamental information about the dataset constructed by various users using our VC-Agent. “Double-Check” refers to the number of times the system initiates the user-assisted double-check process. “Est. Manual Time” denotes the estimated manual collection time. “Actual Time” indicates the actual collection time using our agent (including both user interaction and final automated processing).
}

\label{tab:collected_data}
\end{table*}

%% file: img_table_tex/tab_us_res_quant.tex
\begin{table}[t]
\begin{minipage}[t]{0.45\linewidth}
\vspace{0pt} 
\resizebox{1\linewidth}{!}{%
\begin{tabular}{c|cccc}
\hline
                         & $clip$         & $hpsv2$         & $tf$  & $sc$ \\ \hline
origin\_CogVideoX             & 0.330          & 0.249          & 0.966                & 0.945                \\ 
finetune\_LLAVA-OneVision-7B         & 0.317       &   0.248       &      0.983           &      0.948           \\ 
\textbf{finetune\_VC-Agent (Ours)}   & \textbf{0.344} & \textbf{0.253} & \textbf{0.984}       & \textbf{0.949}           \\ \hline
\end{tabular}%
}
\caption{Quantitative results for T2V generation. “finetune\_” denotes finetuned using the data collected by LLAVA-OneVision-7B or our agent. $tf$ means temporal\_flickering, and $sc$ denotes subject\_consistency.
}
\label{tab:tv_res}
\end{minipage}
\hfill
\begin{minipage}[t]{0.45\linewidth}
\vspace{0pt} 
\resizebox{1\linewidth}{!}{%
\begin{tabular}{l|llllll}
\hline
                  & AP             & AP.5           & AP.75          & AR             & AR.5           & AR.75          \\ \hline
\textbf{finetune} & \textbf{0.682} & \textbf{0.913} & \textbf{0.727} & \textbf{0.757} & \textbf{0.932} & \textbf{0.799} \\
original          & 0.256          & 0.520          & 0.210          & 0.313          & 0.578          & 0.286          \\ \hline
\end{tabular}%
}
\caption{Quantitative results for biped cartoon pose estimation. “finetune” denotes finetuned using the data collected by our agent.}
\label{tab:pose_res}
\end{minipage}
% \vspace{-0.6cm}
\end{table}

%% file: img_table_tex/fig_us_visual_sample.tex
\begin{figure}[t]
    \centering
    \begin{minipage}[t]{0.52\linewidth}
        \vspace{0pt} 
        \centering
        \includegraphics[width=0.9\linewidth]{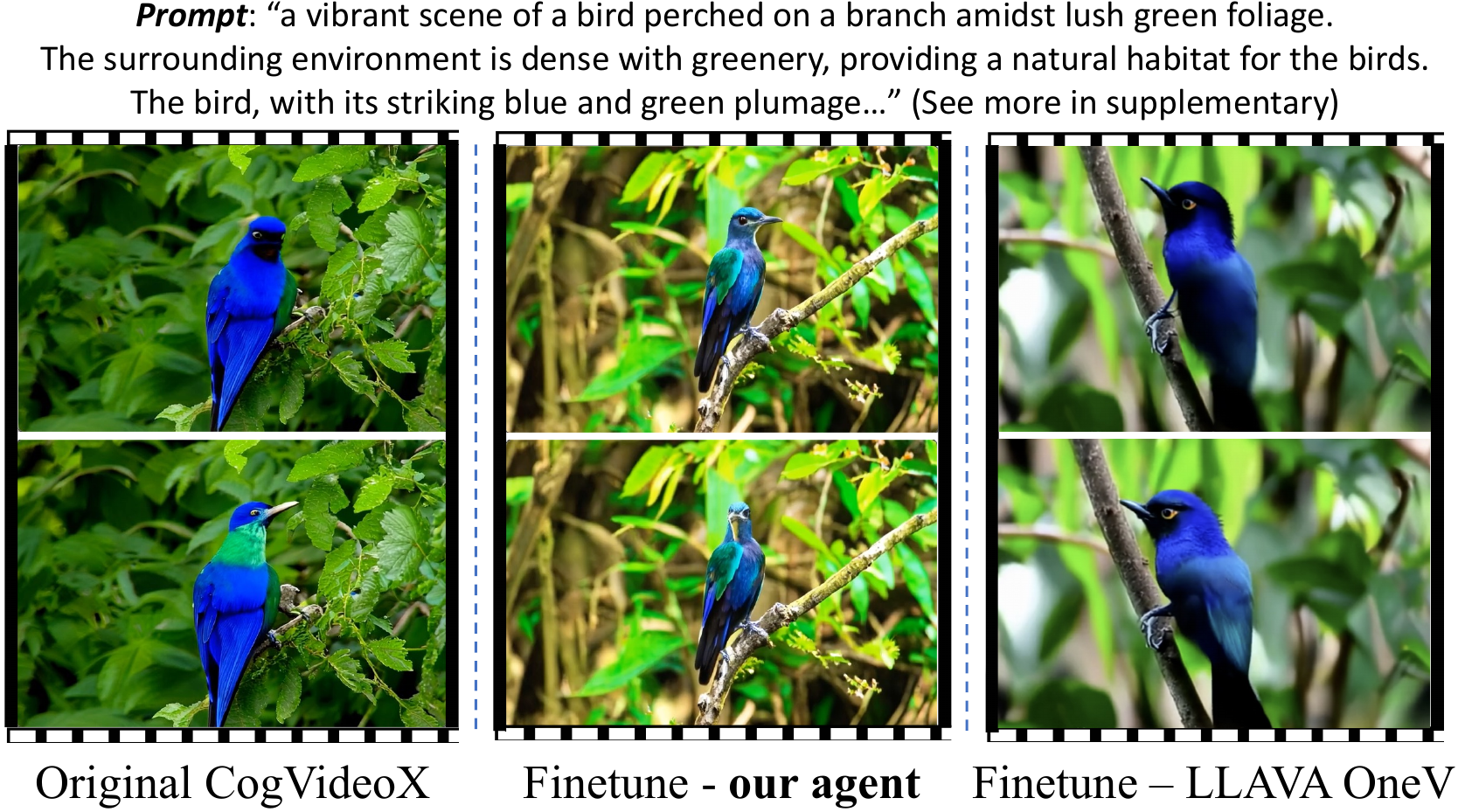}
        \caption{Qualitative results for T2V generation. “Finetune” denotes finetuned using the data from our agent or LLAVA-OneVision. See more in \cref{fig:t2v_extra}.}
        \label{fig:t2v_us2}
    \end{minipage}
    \hfill
    \begin{minipage}[t]{0.42\linewidth}
        \vspace{0pt} 
        \centering
        \includegraphics[width=0.8\linewidth]{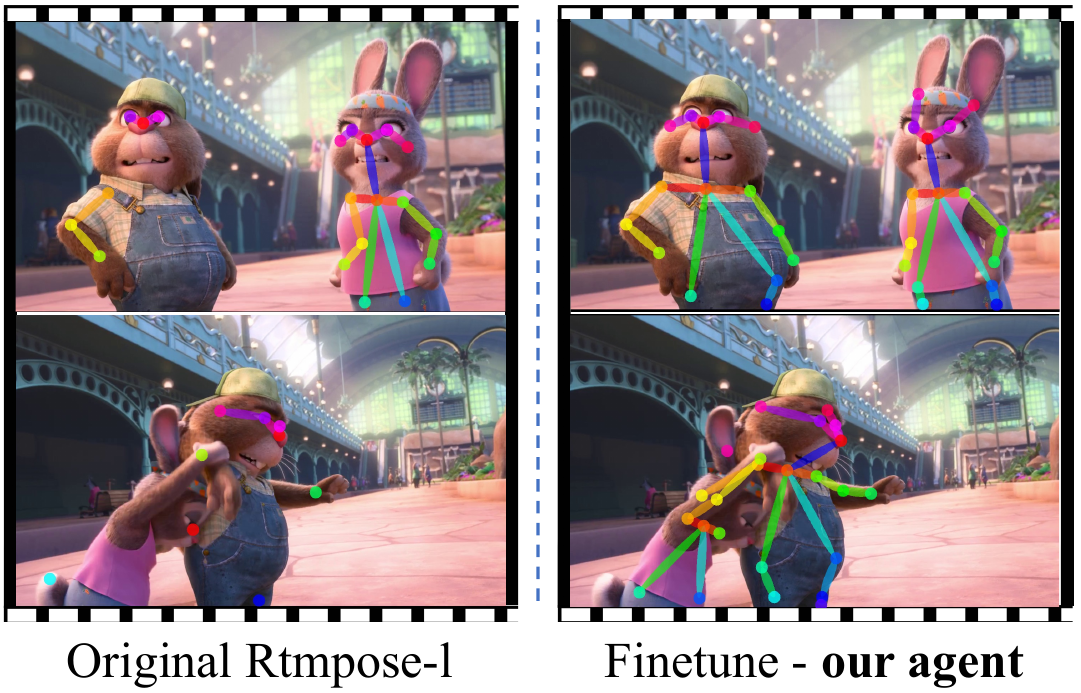}
        \caption{Qualitative results for biped cartoon pose estimation. “Finetune” denotes finetuned using the data from our agent. See more in \cref{fig:pose_extra}.}
        \label{fig:sub}
    \end{minipage}
\end{figure}

%% file: img_table_tex/fig_abl_rounds.tex
\begin{figure}[t]
    \centering
    \begin{minipage}[t]{0.45\linewidth}
        \vspace{0pt}
        \centering
        \includegraphics[width=0.8\linewidth]{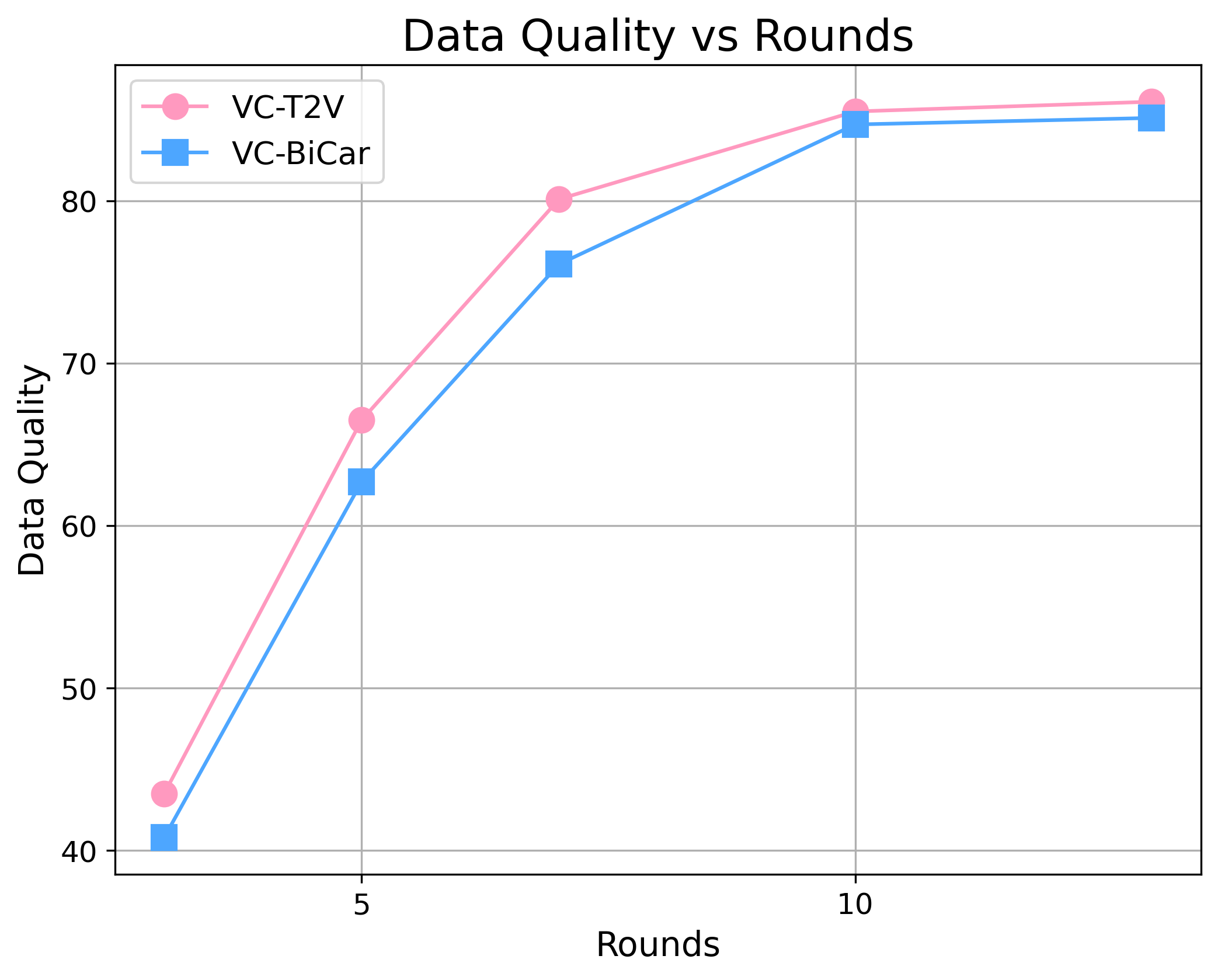}
        % \vspace{-0.3cm}
    \end{minipage}
    \hfill 
    \begin{minipage}[t]{0.45\linewidth}
        \vspace{0pt}
        \centering
        \includegraphics[width=0.8\linewidth]{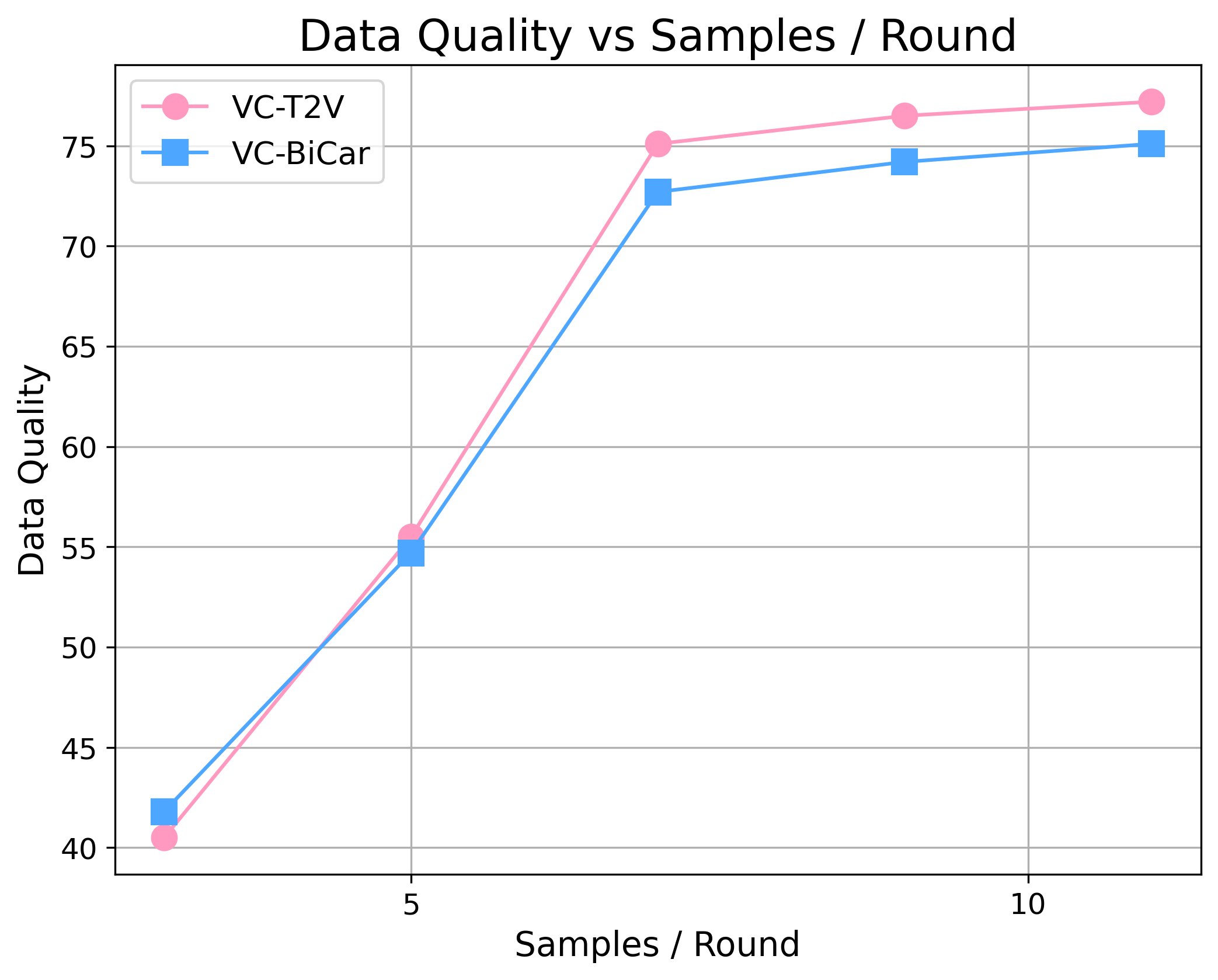}
    \end{minipage}
    \caption{Follow-up survey about “\textit{More user effort, better performance?}”. \textbf{Left}: Data quality under varying numbers of interaction rounds. \textbf{(Right)}: Data quality under different numbers of video samples viewed per round.}
    \label{fig:abl_round}
\end{figure}

%% file: sec/7_conclusion.tex
\section{Conclusion}
We have introduced VC-Agent, a groundbreaking interactive agent that efficiently scales up the customized video dataset collection. Our meticulously designed pipeline leverages existing MLLM and LLM, incorporating a framework that uses text descriptions for video clip proposals and a dual-policy mechanism for effective video filtering. Extensive experiments show that our agent drastically reduces the time users spend while ensuring high-quality outputs, with each dataset typically requiring only a few minutes of user interaction. Our agent has facilitated the creation of 10 large-scale video datasets, which not only push the boundaries of customized dataset construction but also offer a powerful tool across diverse domains.

\begin{figure*}[htbp]
    \centering
    \includegraphics[width=1.0\textwidth]{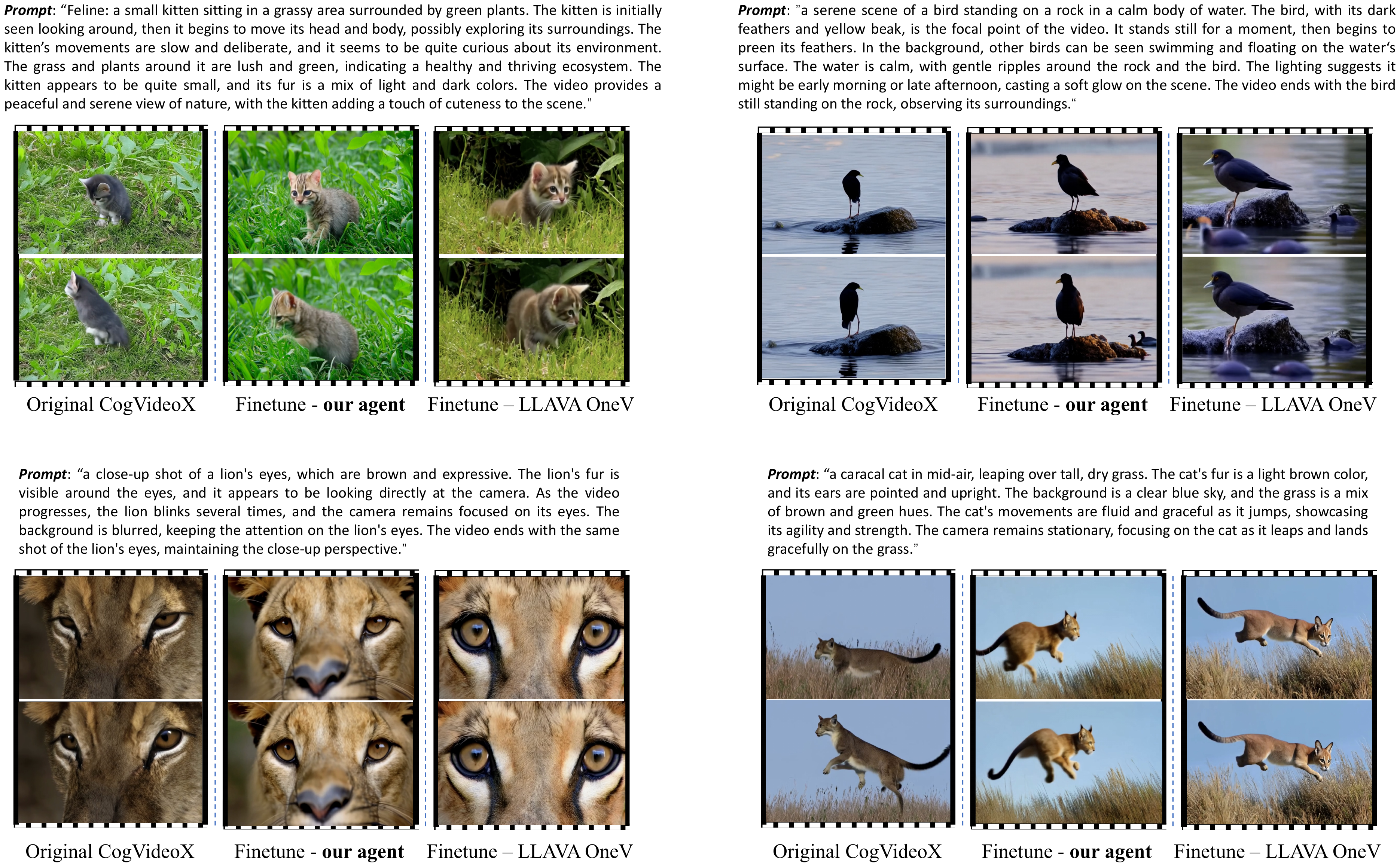}
    \caption{More qualitative results collected from users for text-to-video generation. “Finetune” denotes finetuned
using the collected data from our agent or LLAVA-OneVision-7B \cite{li2024llava}.}
    \label{fig:t2v_extra}
\end{figure*}

\begin{figure*}[htbp]
    \centering
    \includegraphics[width=0.7\textwidth]{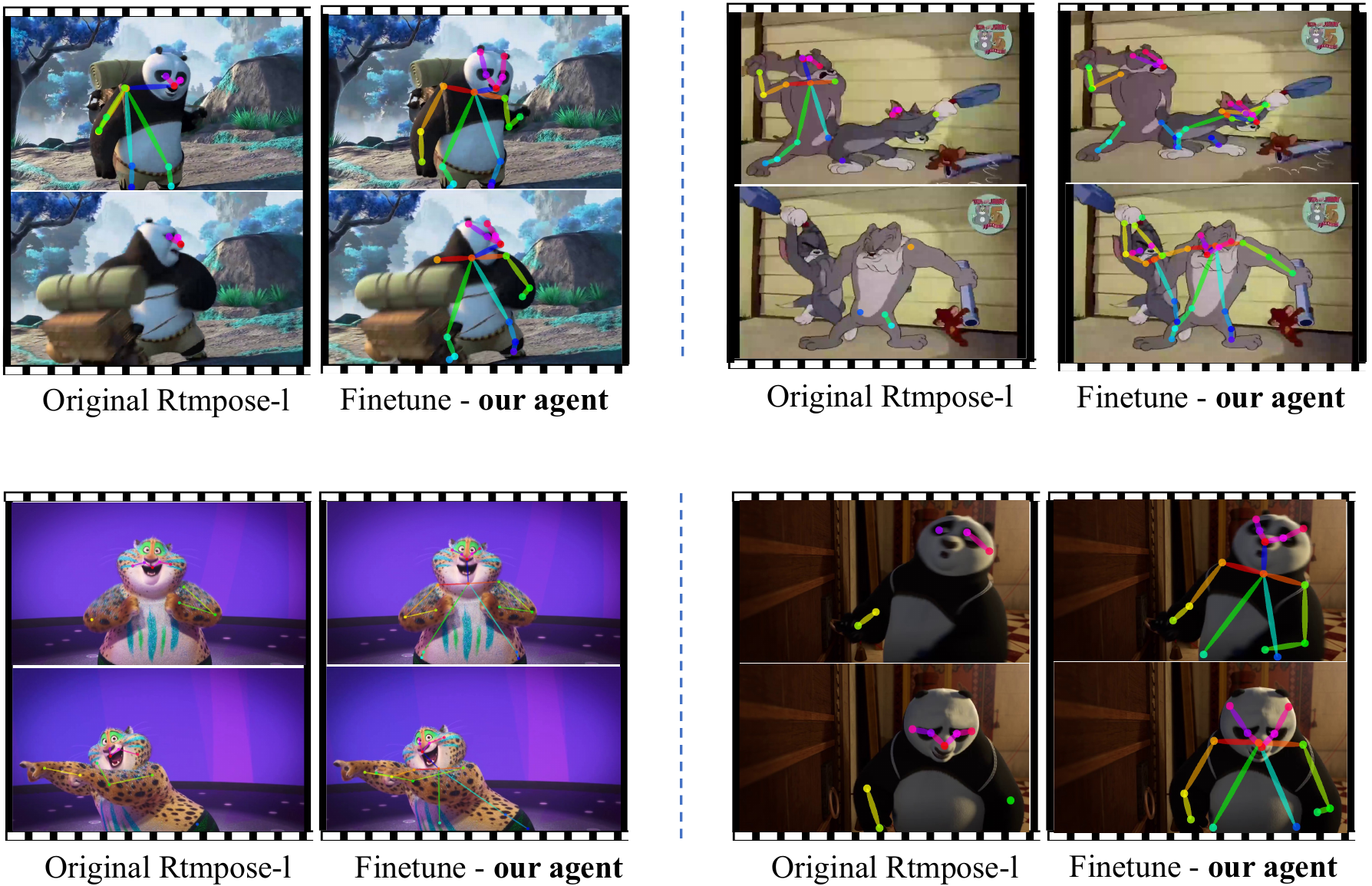}
    \caption{More qualitative results collected from users for biped cartoon pose estimation. “Finetune” denotes finetuned
using the collected data from our agent.}
    \label{fig:pose_extra}
\end{figure*}

\clearpage

%% file: sec/X_suppl.tex
\clearpage
\setcounter{page}{1}

\section{User Study}

\subsection{Questionnaire}
\paragraph{System Usability Scale (SUS):} 
\begin{itemize}
\item I think that I would like to use this system frequently.
\item I found the system unnecessarily complex.
\item I thought the system was easy to use.
\item I think that I would need the support of a technical person to be able to use this system.
\item I found the various functions in this system were well integrated.
\item I thought there was too much inconsistency in this system.
\item I would imagine that most people would learn to use this system very quickly.
\item I found the system very cumbersome to use.
\item I felt very confident using the system.
\item I needed to learn a lot of things before I could get going with this system.
\end{itemize}

\paragraph{Self-Designed Questionnaire:} 
\begin{itemize}
\item What is the purpose task for the dataset you have collected?
\item During the collection process, how many rounds of interaction did you have with the VC-Agent in total?
\item What are the general criteria you have for screening the videos?
\item How much data did you ultimately collect?
\item What percentage of the collected data meets your requirements?
\item Based on your previous experience collecting similar video datasets, how long do you think it would take to collect a dataset of this scale?
\item How many hours did you actually spend building this dataset?
\item Do you have any additional suggestions or feedback?
\item Please give your dataset a name starting with "VC-."
\end{itemize}

\subsection{Questionnaire Results}
\paragraph{Result of SUS}

\input{img_table_tex/fig_sus_res}

\cref{fig:sus_res} illustrates the average scores from the eight questionnaires completed by participants. The design of the SUS questions alternates between positive and negative statements, with scoring ranging from 1 to 5. All participants rated the VC-Agent relatively high, indicating strong usability and satisfaction with our framework. \\

\paragraph{User feedbacks.}
We received rich user feedback from participants, including:
\begin{itemize}
\item  \textbf{``Pretty easy to use; before this, the manual collection process had me feeling dizzy"}(P2)
\item \textbf{``The quality of videos collected by VC-Agent significantly exceeded my expectations, and the speed of final automatic collection was remarkably fast, tens of thousands of videos had been collected after a sleep."}(P4)
\item \textbf{``Sometimes the system will ask me to double-check a specific attribute for some interesting samples. This function is really useful, as it prompts me to clear my requirements more specifically."}(P3)
\item \textbf{``The process of interacting with VC-Agent enabled me to clarify my dataset requirements incrementally, and the filter requirements became a kind of pseudo caption."}(P7) 
\item \textbf{``Normally, just after about 10 interactions, the returned samples became stably satisfying."}(P6)
\item \textbf{``For some complex requests, VC-Agent may not perform particularly well, such as `I need videos captured with a multi-view surrounding shot.'"}(P5)
\item \textbf{``When there are too many filtering criteria, the speed will be slightly slower."}(P8).
\end{itemize}

\subsection{Follow-up Survey}

\paragraph{Text-to-video prompts.}

\begin{itemize}
\item \textit{Bird:} “a vibrant scene of a bird perched on a branch amidst lush green foliage. The bird, with its striking blue and green plumage, remains motionless for a brief moment before it begins to preen itself. The surrounding environment is dense with greenery, providing a natural habitat for the birds. The camera remains stationary, allowing viewers to appreciate the intricate details of the bird's feathers and the intricate patterns of the leaves in the background. The video provides a serene glimpse into the life of this beautiful bird in its natural setting.”
\end{itemize}

\paragraph{User requirements.}

Major data set requirements for text-to-video generation and estimation of the pose of biped cartoon characters were offered by the two participants.
\begin{itemize}
\item Animal Video
\item Cat-like Animal
\item Close-up shot
\item Single object in the video
\item The object should not be obscured 
\item The full body should be shown in the video.
\end{itemize}

\begin{itemize}
\item Animation Style Video
\item Less than three characters in film
\item Bipedal Humanoid characters that have the human pose
\item Close-up shot
\item The animated character needs to be anthropomorphized and have a human-like posture. 
\item Most parts of the body should be in the film.
\end{itemize}

\section{More Implementation Details and Demo}
\paragraph{Web Crawler}
We deploy the crawler based on MediaCrawler \cite{MediaCrawler2024} for retrieving the video information and URLs from the widely used video websites. The searching keywords are provided and updated by our agent during the user interaction phase. After the video information has been collected, a video download script based on Lux \cite{Lux2024} will download the raw data during the video proposal phase.

\paragraph{Web UI}
We provide a demo video of the whole operation of our agent in the Web UI. You may refer to the supplementary file folder and access the video file named \textcolor{red}{\textit{ui\_demo.mp4}}.
In this video, the user starts to build a new project, typing the initial query at 00:13. After getting the videos returned by the agent, the user checks the videos, conducts confirmation and provides comments, ending at 01:17. Next, the agent will analyze our requirements, create/update the filtering policy, and return a new batch of filtered videos. This interaction is repeated for multiple rounds until the user accepts all the returned videos at 03:00, prompting our agent to use the final updated policy to collect data in a fully automatic manner. The VC-Agent would continuously collect videos in the backstage server and filter videos according to the policy. Finally, the output videos are shown (from 03:15 to the end).

\paragraph{Storage and computing resources.}
We deploy the LLM and MLLM model based on the framework of vllm \cite{kwon2023efficientmemorymanagementlarge}. For each of the large language models, we use 1 NVIDIA A100-80G GPU for development. To store massive amounts of online video data, we have allocated 20TB of solid-state drives for video storage. Additionally, we are deploying cloud storage space to further enhance the storage and the quality of the user experience. 

\paragraph{Handle inconsistent feedback.}
Inconsistent or uninformative feedback from the user on a few examples (“noisy” labels) has a negligible impact, yet frequent inaccuracies may degrade performance. As such, and considering our interaction is time-efficient, we recommend resetting when users substantially change ideas.

\section{More Results and Ablation Studies}

\paragraph{More qualitative benchmark results.}
\cref{fig:abl_effect} compares more qualitative benchmark results of progressively adding filtering requirements, showing the effectiveness of our method.

\input{img_table_tex/supp/fig_progress_sup}

\input{img_table_tex/abl_mllm}

\paragraph{Backbones.}
The ablation of using different MLLM backbones is shown in \cref{tab:abl_mllm}. We use LLava-OneVision \cite{li2024llava} as our default MLLM backbone.

\section{What kind of video properties can be handled?}
Metadata is often available in video captions (can be directly crawled), while our work focuses on non-captioned information.
Moreover, according to user feedback and our testing, our framework can handle general cases of video subtitles, types, dynamic/static, whether synthetically generated (R.3, third domain of our benchmark).

However, it does not perform well when handling some corner/extreme cases. As illustrated in \cref{fig:failure}, our agent struggles to accurately distinguish \textit{highly} realistic virtual scenes (\eg, the examples on the right in the first and second rows). Additionally, our method exhibits difficulty in reliably recognizing \textit{transient} object movements, such as hands and subtitles, that suddenly appear in the frame (\eg, the two examples in the third row). For some \textit{action/semantic-ambiguous} cases, our framework also makes errors, \eg, in the left image of the first row, although a basketball player kicks a basketball on the court, it still classifies the activity as playing basketball.

Moreover, our method can not effectively process some other types of complex user requirements, such as cut detection, motion degree, and camera movement/trajectory (\eg, when users collect multi-view surround videos).

Nevertheless, we think that these abilities depend largely on MLLMs, which isn't our core focus - we focus on providing an \textbf{interactive system}, where MLLMs serve as the \textit{parsing tool}.

\input{img_table_tex/supp/fig_failure_case}

\section{Benchmark Statistics}

\input{img_table_tex/tab_pvb_distribution}

The statistics of our proposed Personalized Video Collection Benchmark are provided in \cref{tb:benchmark1_st}. We quantify the number of videos that meet the first requirement (R.1) throughout to meet all five requirements (R.5). For example, R.1\&2 denotes that the video meets both R.1 and R.2. The purpose of this split is to assess the model's ability to accurately collect videos that meet requirements from \textit{varying extent of specification}.

\section{Prompt}

\paragraph{Grounding keywords.}
Your task is to help the user generate the prompt for a grounding detection network. You will get the asking and comments from a user who wants to build a certain type of video dataset; the first message from the user would be the demand for the video dataset, and the following messages are the comments for the videos the user has already seen. You should summarize the user demand and generate the prompt for the grounding module to detect the correct part of the video as the user will. Try to use short phrases, including the most important keywords you include from the user inquiry. The output phrase should follow the format: [GRN] phrase [/GRN]

\paragraph{Crawler keywords}
You are helping the user build the video dataset. Your duty is to generate the search keywords based on user demand, which can help the website crawler search for related videos on the Chinese video website. The keywords should follow the format: [KEY] keyword1, keyword2, keyword3 [/KEY] 

\paragraph{User demand summarization.}
You are an agent for assisting the user in building a specific kind of video dataset.  You need to grasp the core demands of the users. You will also receive many user comments on the video in the future, and you need to summarize the important information from user comments into user requirements. The output requirements text should follow the format: [TXT] requirements text [/TXT]

\paragraph{Description summarization.}
Summarize the description in the following, and extract the most important information related to [FEATURE]. Your summarization should be in one paragraph.

\paragraph{Video descriptor.}

Task: Analyze the video, and you will be given a prompt indicating the requirements for the video. Based on the video content, analyze whether the video meets the requirement or not, answer the question, and give the related content description in the video. Answer the video meets the requirement or not, and give a yes or no answer strictly as follows:
1. Yes/No: Answer the question with Yes or No. 
2. Evidence: Provide a brief explanation about the correlated evidence in the video for your answer.
3. Summary: Provide a concise conclusion about the presence or absence of the feature.
Ensure your explanation is clear, accurate, and directly addresses the task. [QST] Question [/QST]

\section{Limitations.} 
One limitation of our current approach is that, as an early exploration, there are still some challenges in the filtering process, particularly when handling more complex user requirements. Moreover, there is still room for improvement in both efficiency and accuracy for the VC-Agent. 

Additionally, it still requires human confirmation and judgment regarding the quality of data collection for more complex and intricate task requirements.

%% file: img_table_tex/fig_sus_res.tex
\begin{figure}[ht]
    \centering
    \includegraphics[width=0.5\textwidth]{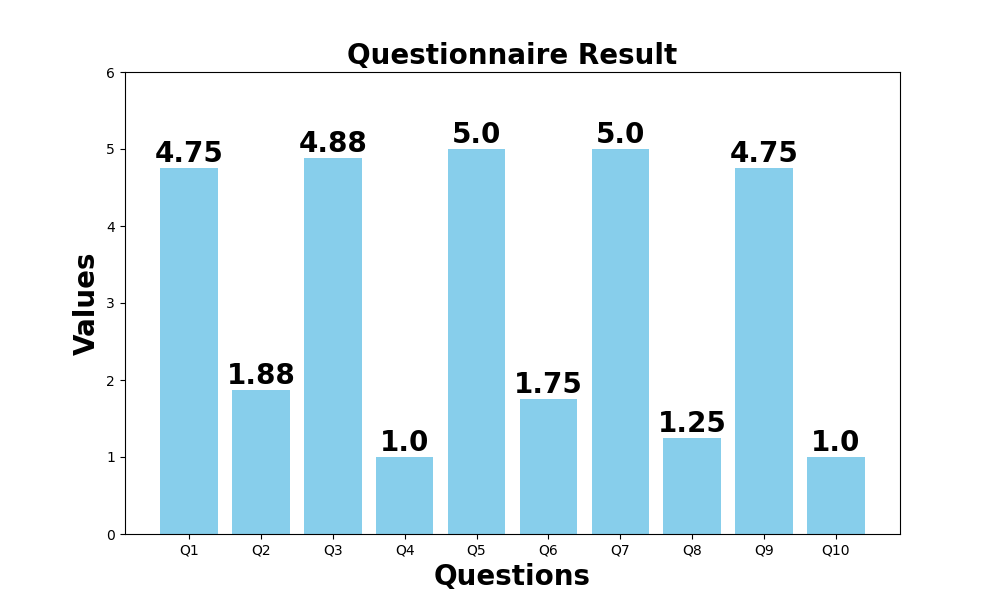} % 替换为你的图片文件名
    \label{fig:sus}
    \caption{The results of the SUS questionnaire. The design of the SUS questions alternates between positive and negative questions. The survey results show that users provided very positive feedback regarding the VC-Agent in terms of effectiveness, efficiency, and satisfaction. }
    \label{fig:sus_res}
\end{figure}

%% file: img_table_tex/supp/fig_progress_sup.tex
\begin{figure}[t]
    \centering
    \includegraphics[width=1.0\linewidth]{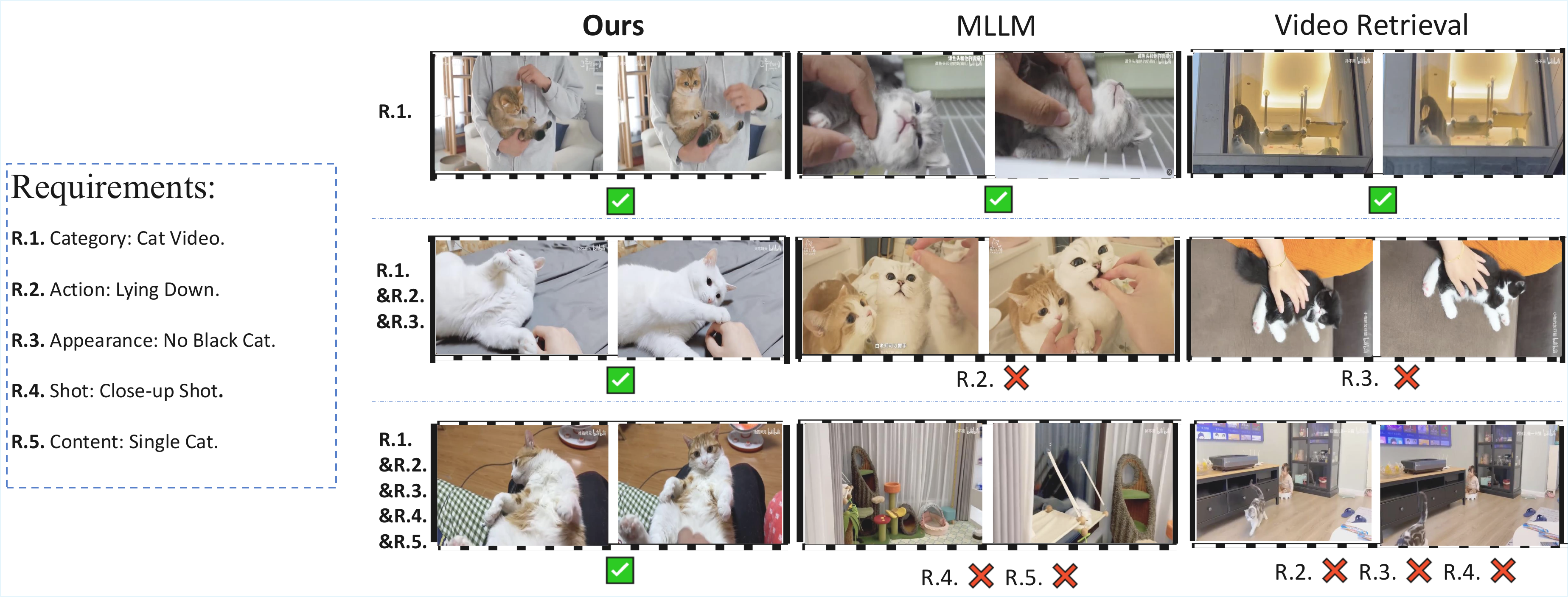} 
    % \vspace{-0.2cm}
    \caption{More filtering results obtained by incrementally adding different requirements. “Ours” denotes filtering conducted using the VC-Agent, “MLLM” refers to LLAVA-OneVision \cite{li2024llava}, and “Video Retrieval” represents GRAM \cite{cicchetti2024gramian}.}
    
    \label{fig:progress_sup}
    % \vspace{-0.3cm}
\end{figure}

%% file: img_table_tex/abl_mllm.tex
\begin{table}[]
\centering
\resizebox{0.5\linewidth}{!}{%
\begin{tabular}{l|lll}
\hline
Requirements     & R.1   & R.1\&2\&3 & R.1\&2\&3\&4\&5 \\ \hline
MiniGpt4-Video   & 52.32 & 47.18     & 40.83           \\
LLAVA-Next-Video & 61.12 & 53.14     & 48.65           \\
LLAVA-OneVision  & 64.82 & 56.23     & 49.17           \\ \hline
\end{tabular}%
}
\caption{The results of the ablation experiments for the base model of the VC-Agent}
\label{tab:abl_mllm}
\vspace{-0.72cm}
\end{table}

%% file: img_table_tex/supp/fig_failure_case.tex
\begin{figure}[t]
    \centering
    \includegraphics[width=1.0\linewidth]{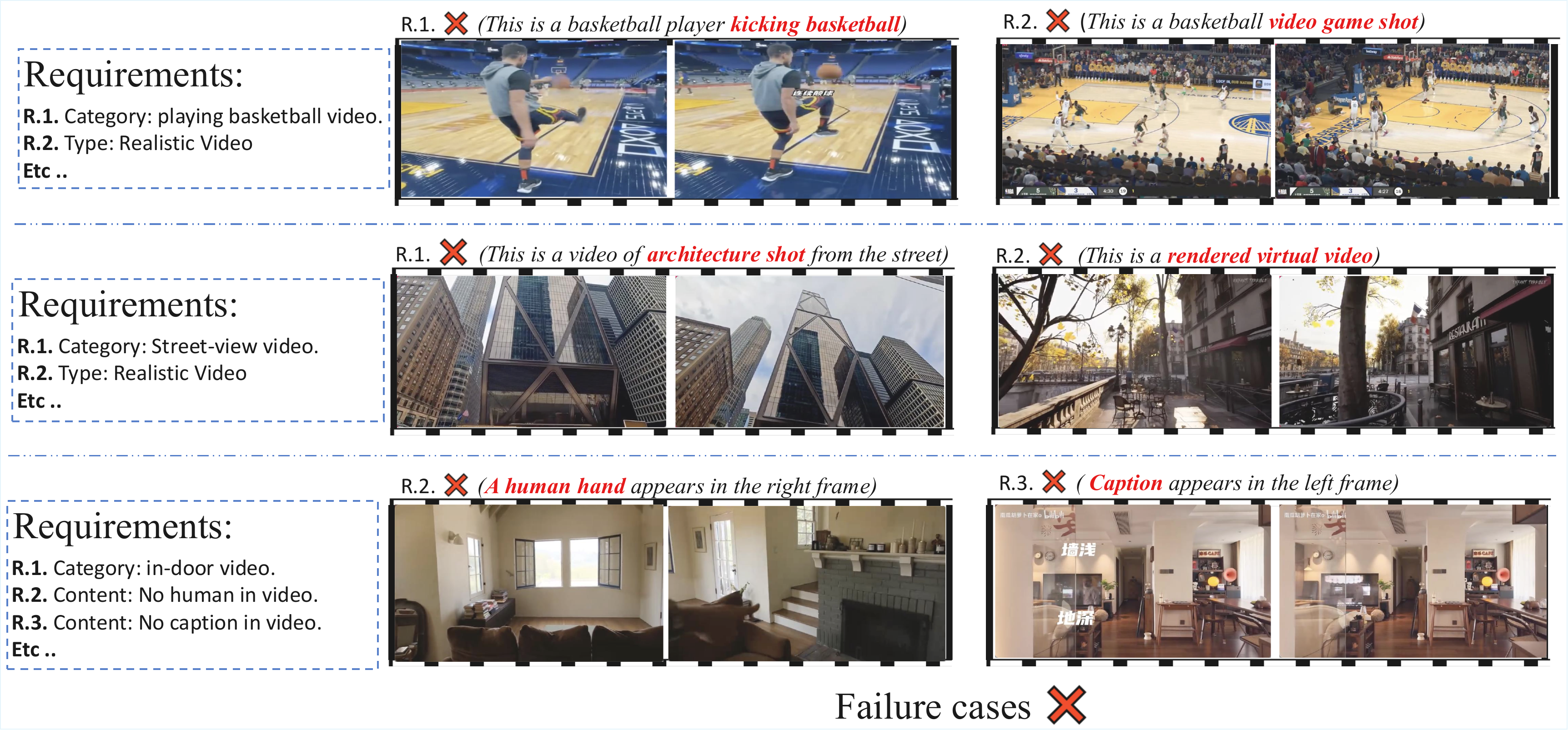} 
    % \vspace{-0.2cm}
    \caption{Some failure cases of our agent observed during user studies.}
    \label{fig:failure}
    \vspace{-0.7cm}
\end{figure}

%% file: img_table_tex/tab_pvb_distribution.tex
\begin{table}[t]
\centering
\resizebox{0.6\linewidth}{!}{
\begin{tabular}{l|lllll}
\hline
Requirements & R.1    & R.1\&2    & R.1\&2\&3    & R.1\&2\&3\&4    & R.1\&2\&3\&4\&5   \\ \hline
Domain 1 &1601  &1342 &1207  & 705  & 620 \\ 
Domain 2 &952   &605  &533   & 487   & 415 \\ 
Domain 3 &1009  &683  &591   & 388   & 181 \\  \hline
Total        & 3562 & 2630 & 2331 & 1580 & 1316   \\ \hline
\end{tabular}
}
\caption{The statistics of our Personalized Video Collection Benchmark. R.n indicates the n-th requirement that the video needs to meet. R.1\&2 denotes both R.1 and R.2 need to be satisfied, and so on.}
\label{tb:benchmark1_st}
\vspace{-0.8cm}
\end{table}